\documentclass{article} 
\usepackage{iclr2026_conference,times}
\usepackage[T1]{fontenc}
\usepackage{booktabs}
\usepackage{geometry}
\usepackage{multirow}

\usepackage{amsmath,amsfonts,bm}

\def\eqref#1{equation~\ref{#1}}

\def\1{\bm{1}}

\DeclareMathAlphabet{\mathsfit}{\encodingdefault}{\sfdefault}{m}{sl}
\SetMathAlphabet{\mathsfit}{bold}{\encodingdefault}{\sfdefault}{bx}{n}

\makeatletter
\def\iclrheading#1{}
\makeatother

\usepackage{hyperref}
\usepackage{url}
\usepackage{epigraph}
\usepackage{amsmath}
\usepackage{amsfonts}
\usepackage{graphicx}
\usepackage{xcolor}
\usepackage{tcolorbox}
\usepackage[labelformat=simple]{subcaption}

\usepackage[export]{adjustbox}
\usepackage{calc}
\usepackage{amsthm}      
\usepackage[framemethod=default]{mdframed} 

\newcommand{\change}[2]{{#2}}

\newtheoremstyle{compact_style}
  {0.5ex} 
  {0.5ex} 
  {\itshape} 
  {} 
  {\bfseries} 
  {.} 
  {.5em} 
  {} 

\theoremstyle{compact_style}

\surroundwithmdframed[
  linewidth=0.5pt,       
  topline=true,          
  bottomline=true,
  leftline=true,
  rightline=true,
  innerleftmargin=8pt,   
  innerrightmargin=8pt,
  innertopmargin=8pt,
  innerbottommargin=8pt,
]{hypothesis}

\iclrfinalcopy
\title{Emergent Slow Thinking in LLMs as Inverse Tree Freezing}

\author{Sihan Hu$^{1}$\thanks{These authors contributed equally to this paper.}, Xiansheng Cai$^{3}$$^{*}$, Yuan Huang$^{5}$\thanks{Corresponding Author}, Zhiyuan Yao$^{6}$\thanks{Corresponding Author}, Linfeng Zhang$^{5,7}$\thanks{Corresponding Author}, Pan Zhang$^{3,4,1}$\thanks{Corresponding Author},\\
\textbf{Youjin Deng}$^{1,2}$\thanks{Corresponding Author}, \textbf{Kun Chen}$^{3}$\thanks{Corresponding Author} \\
$^{1}$Hefei National Laboratory, University of Science and Technology of China \\
$^{2}$Hefei National Laboratory for Physical Sciences at the Microscale and Department of Modern Physics, \\ University of Science and Technology of China\\
$^{3}$Institute of Theoretical Physics, Chinese Academy of Sciences\\
$^{4}$School of Fundamental Physics and Mathematical Sciences, Hangzhou Institute for Advanced Study, \\UCAS \\
$^{5}$DP Technology\\
$^{6}$Lanzhou Center for Theoretical Physics, Key Laboratory of Theoretical Physics of Gansu Province, \\
 Key Laboratory of Quantum Theory and Applications of MoE,\\
 Gansu Provincial Research Center for Basic Disciplines of Quantum Physics, Lanzhou University \\
$^{7}$AI for Science Institute, Beijing\\
\texttt{huangyuan@dp.tech, yaozy@lzu.edu.cn, zhanglf@dp.tech,}\\
\texttt{panzhang@itp.ac.cn, yjdeng@ustc.edu.cn, chenkun@itp.ac.cn}
}

\graphicspath{{figures/}}

\usepackage[section]{placeins}

\begin{document}
\maketitle
\begin{abstract}
Reinforcement learning with verifiable rewards (RLVR) enables
large language models to acquire slow, multi-step reasoning from sparse
final-answer signals. We provide a statistical-physics picture of this emergence.
We show that an autoregressive model's finite capacity forces it to compress
its exponentially large prefix space into a Markov network of predictive states,
on which slow thinking unfolds as a random walk---the Concept Network
(CoNet) picture. Within CoNet, RLVR dynamics are governed by two mechanisms:
merging of compatible paths and frustrated competition among incompatible ones.
Together they drive the network through nucleation, growth, and freezing into
multi-input, single-output directed inverse trees. The picture reproduces the
training dynamics of a 1.5-billion-parameter LLM and yields three predictions:
reasoning chains lengthen as a geometric necessity of sparse topology; SFT
induces catastrophic forgetting through bridge-node rupture;
and frustration drives policy collapse. Building on the structural
timing inherent in inverse-tree freezing, we propose Annealed-RLVR---a brief SFT
intervention at the moment of maximum frustration. It outperforms standard
RLVR on both in- and out-of-distribution benchmarks, with the largest gains at
high sampling budgets where standard RLVR collapses. The same SFT applied after
the trees freeze instead triggers catastrophic forgetting, isolating timing as
the active ingredient.
\end{abstract}

\section{Introduction}\label{sec:intro}

Cognitive science draws a sharp distinction between fast thinking (System~1) and slow thinking (System~2)~\citep{kahneman2011thinking,yang2026slowthinking}: the former is rapid, automatic, pattern-driven intuition; the latter is slow, deliberate, step-by-step rational reasoning. A large language model (LLM)---an autoregressive sequence generator built on the Transformer architecture~\citep{vaswani2017attention}---is, by construction, a System~1 machine~\citep{hua-zhang-2022-system}: given a token sequence $(a_1, a_2, \ldots, a_t)$, the model uses its parameters $\theta$ to emit, in a single forward pass, the conditional probability $\pi_\theta(a_{t+1}\,|\,a_1,\ldots,a_t)$ of the next token; the sampled $a_{t+1}$ is appended to the sequence and generation continues, with no explicit search or planning at any point. A recent finding has overturned this picture: an extraordinarily simple training paradigm---reinforcement learning with verifiable rewards (RLVR)~\citep{drgrpo,deepseek-r1,teamKimiK15Scaling2025,qwen3}---suffices for LLMs to spontaneously acquire slow thinking~\citep{yang2026slowthinking}. After the model produces a complete response, RLVR issues a sparse $\{0,1\}$ reward based solely on the correctness of the final answer, with no supervision of intermediate steps, and the parameters $\theta$ are then updated by policy gradient. From GPT-o1~\citep{jaech2024openai} to DeepSeek-R1~\citep{deepseek-r1}, RLVR-trained models spontaneously unfold reasoning chains hundreds of thousands of tokens long---decomposing problems step by step, checking intermediate results, correcting errors---reaching unprecedented levels in mathematical proof and program generation~\citep{team2026kimi,zeng2026glm,singh2025openai,deepseekai2026deepseekv4}. The crux is that the model is never taught \emph{how} to reason---it is told only whether the answer is right---and slow thinking emerges spontaneously from the fast-thinking substrate.

From the standpoint of statistical physics, the emergence of slow thinking on top of a fast-thinking substrate carries the classical hallmark of an \emph{emergent} phenomenon~\citep{anderson1972more,cai2025learning}: a separation of timescales, with microscopic degrees of freedom (individual tokens and their local statistics) unchanged yet a wholly new collective behaviour (multi-step logical reasoning) appearing at the macroscopic level. Wang \textit{et al.}~\citep{wang2025beyond} provide direct empirical evidence for this separation through token-by-token information-entropy measurements (the model's uncertainty over the next token) of generated sequences before and after RLVR training: about 80\% of tokens lie within local computation chunks (multi-token phrases with near-deterministic transitions) where the entropy is very low ($<0.5$\,nats) and the probability distribution barely changes between pre- and post-RLVR---these are the fast computation steps already mastered during pretraining; what RLVR genuinely reshapes is the remaining ${\sim}20\%$ of high-entropy tokens located at the transitions between chunks, which dictate the macroscopic trajectory of the reasoning chain---exactly where slow thinking takes place. Even more telling, updating only these 20\% high-entropy tokens reproduces the performance gain of full training~\citep{wang2025beyond}. Fast thinking (within chunks) stays put; slow thinking (the joints between chunks) emerges---a clean separation of timescales between the two.

Not every system trained with reinforcement learning develops emergent reasoning. AlphaGo~\citep{silver2016mastering}, trained over the vast search space of Go ($\sim\!10^{170}$ legal positions), never spontaneously acquired a search strategy of its own---search had to be hard-coded into the algorithm via Monte Carlo tree search (MCTS)~\citep{coulom2006efficient,browne2012survey}. The difference lies in two structural features that LLMs possess and AlphaGo does not. The first is a \emph{domain-agnostic action-sequence representation}. AlphaGo's policy takes the current board (a state representation) as input; the move history is discarded, and a search strategy cannot be expressed in the network's input/output format and must be bolted on externally as MCTS. An LLM's policy, by contrast, takes the full token-generation history $(a_1,\ldots,a_t)$ as input (an action-sequence representation). Under this representation, fast and slow thinking share a single format---both are token sequences differing only in length---and a reasoning strategy is itself a pattern expressible and learnable within that sequence, requiring no external search module. The second is \emph{self-reference}: every output $a_{t+1}$ is immediately fed back as part of the input for the next step, contributing to the prediction of $a_{t+2}$. Self-reference equips the system with an additional axis along which it can scale---inference time. Beyond the parameter count $|\theta|$ and the size of training data, it opens a third scaling channel, namely test-time scaling~\citep{jaech2024openai,snell2024scaling,brown2024large}, allowing a policy in the unified action space to deepen itself through stepwise iteration.

These two structural features explain \emph{why} LLMs can develop emergent slow thinking---they possess a domain-general action space and a scalable temporal axis. They have not yet answered a more fundamental question: what does RLVR actually change at the microscopic level so that macroscopic reasoning can emerge?

The first contribution of this work is to introduce, for Transformer-class autoregressive models, an effective Markov representation built on the idea of predictive state variables (PSVs)~\citep{littman2001predictive}. Under this representation, the exponentially large token-sequence space is compressed onto a sparse network of effective states, and reasoning is mapped to a random walk on that network. This framework also resolves a practical puzzle: why does reasoning training in LLMs require a Group Relative Policy Optimization (GRPO) based RLVR~\citep{shao_deepseekmath_2024} while Monte Carlo tree search (MCTS) is unable to do the job? The reason is that MCTS relies on an explicit state representation (such as a board position) to construct its search tree, whereas the effective states of an action-sequence representation only emerge implicitly through training and cannot be enumerated in advance.

Building on this effective Markov representation, we propose that the training dynamics of RLVR can be mapped onto a freezing problem on a complex network. Multi-task RLVR with shared parameters merges paths and carves the high-probability edges into a sparse, multi-input, single-output directed tree---an inverse tree. The inverse tree forms spontaneously through processes analogous to nucleation and growth in a supercooled liquid~\citep{avrami1939kinetics,kelton2010nucleation}. Almost every node on the tree has a single, near-deterministic outgoing edge, ensuring that even though reasoning may span many steps, each individual choice is nearly certain and unlikely to err---precisely the structural guarantee that allows long-range reasoning to proceed reliably. At the same time, the reasoning paths of different tasks compete on shared nodes, producing a \emph{frustration} mechanism: when two reasoning trajectories share an intermediate node but demand different outgoing edges, competition at that node eventually freezes the dominant path into a deterministic transition while almost entirely erasing the subordinate one.

This theoretical framework gives a unified account of several core phenomena in RLVR training: the lengthening of reasoning chains with growing capability~\citep{deepscaler2025,skywork-or1-2025} follows naturally from the geometry of a sparse tree (\emph{Inverse Tree Picture} section); the catastrophic forgetting triggered by interleaving SFT into RL~\citep{li2017learning,luo2025empiricalstudycatastrophicforgetting,ding2025improvedsupervisedfinetuninglarge} stems from the rupture of critical branching nodes on the inverse tree (\emph{Catastrophic Forgetting} section); and a proper understanding of the forgetting-by-frustration mechanism (\emph{Two Core Mechanisms} section) points to a favourable timing window for SFT intervention---the Annealed-RLVR scheme (\emph{Annealed-RLVR} section), which suppresses policy collapse~\citep{cui2025entropymechanismreinforcementlearning,yue2025doesreinforcementlearningreally} while preserving high exploration capacity.

\section{Effective Markov Representation and State Compression}\label{sec:markov}

The autoregressive generation of a Transformer is a textbook non-Markov process: the output at step $t$ depends on the full history $(a_1,\ldots,a_t)$ rather than on a finite-dimensional hidden state. This dependence on the entire history is usually counted among the strengths of Transformers over models with explicit state representations such as hidden Markov models---it grants greater expressive power. By the same token, however, the dynamics of an autoregressive model is hard to analyse theoretically: treating each distinct history as a separate state yields a state space of size $V^t$ ($V$ the vocabulary size, $t$ the sequence length), which explodes exponentially in $t$.

The work of Littman, Sutton and Singh on predictive state representations~\citep{littman2001predictive} provides a key insight: for any observable dynamical system, no matter how intricate its internal mechanism, the state can be characterised equivalently by a set of predictions about future observables---the number of effective states is determined by the number of distinguishable future behaviours, not by the size of the history space. For an autoregressive model with $d = |\theta|$ parameters this means that the parameter space can encode at most $O(e^d)$ distinguishable conditional distributions, so a great many distinct histories must necessarily map to identical future behaviour.

Following this line of thought, we define the \emph{predictive state variable} (PSV): given a prompt $Q$ and an action history $h$, two pairs $(Q,h)$ and $(Q',h')$ belong to the same effective state if and only if they induce future conditional distributions that are indistinguishable at coarse-graining scale $\epsilon$,
\begin{equation}\label{eq:psv}
(Q,h) \sim_\epsilon (Q',h') \;\Longleftrightarrow\; \|\pi_\theta(\cdot\,|\,Q,h) - \pi_\theta(\cdot\,|\,Q',h')\| < \epsilon,
\end{equation}
where $\pi_\theta(\cdot\,|\,Q,h)$ denotes not the single-step distribution but the joint conditional distribution over the entire remaining token sequence. The threshold $\epsilon$ has a natural lower bound set by the finite context window and finite vocabulary. By construction, members of the same PSV class $s = [(Q,h)]_\sim$ produce identical future distributions up to $\epsilon$. The next token therefore depends only on $s$, not on the specific history that reached it---the defining property of a Markov state. A crucial property follows: histories from different problems can be merged into a single equivalence class through compression. This cross-problem state sharing will become the microscopic basis of path merging and inverse-tree formation in later sections.

\begin{figure}[!htbp]
\centering
\begin{subfigure}[b]{0.75\textwidth}
\centering
\includegraphics[width=\textwidth]{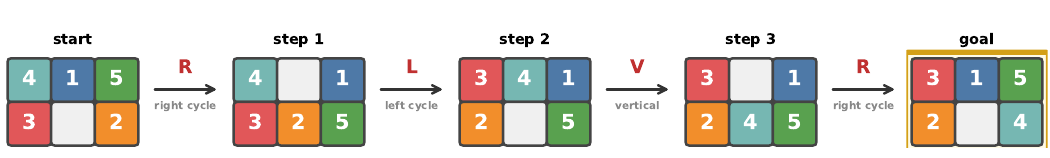}
\caption{}
\end{subfigure}
\vspace{0.3em}\\
\begin{subfigure}[b]{0.45\textwidth}
\centering
\includegraphics[width=\textwidth]{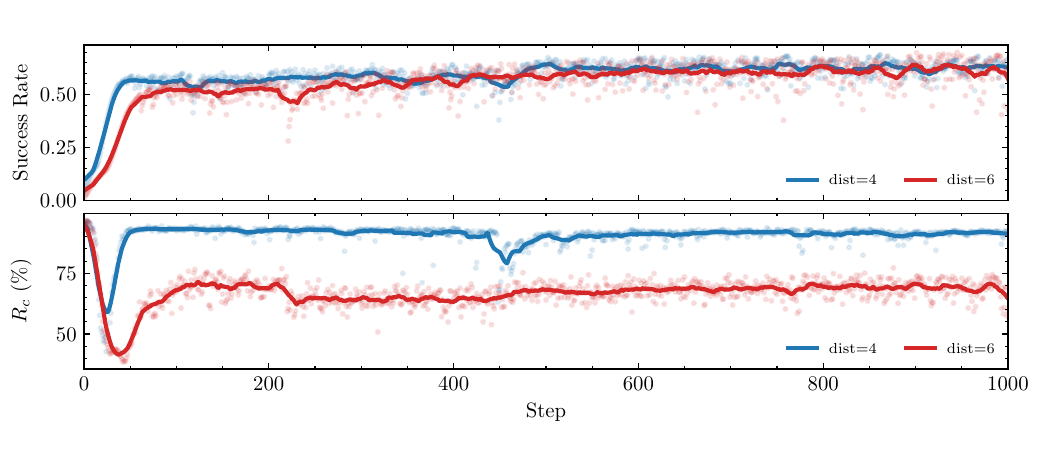}
\caption{}
\end{subfigure}
\hfill
\begin{subfigure}[b]{0.5\textwidth}
\centering
\includegraphics[width=0.48\textwidth]{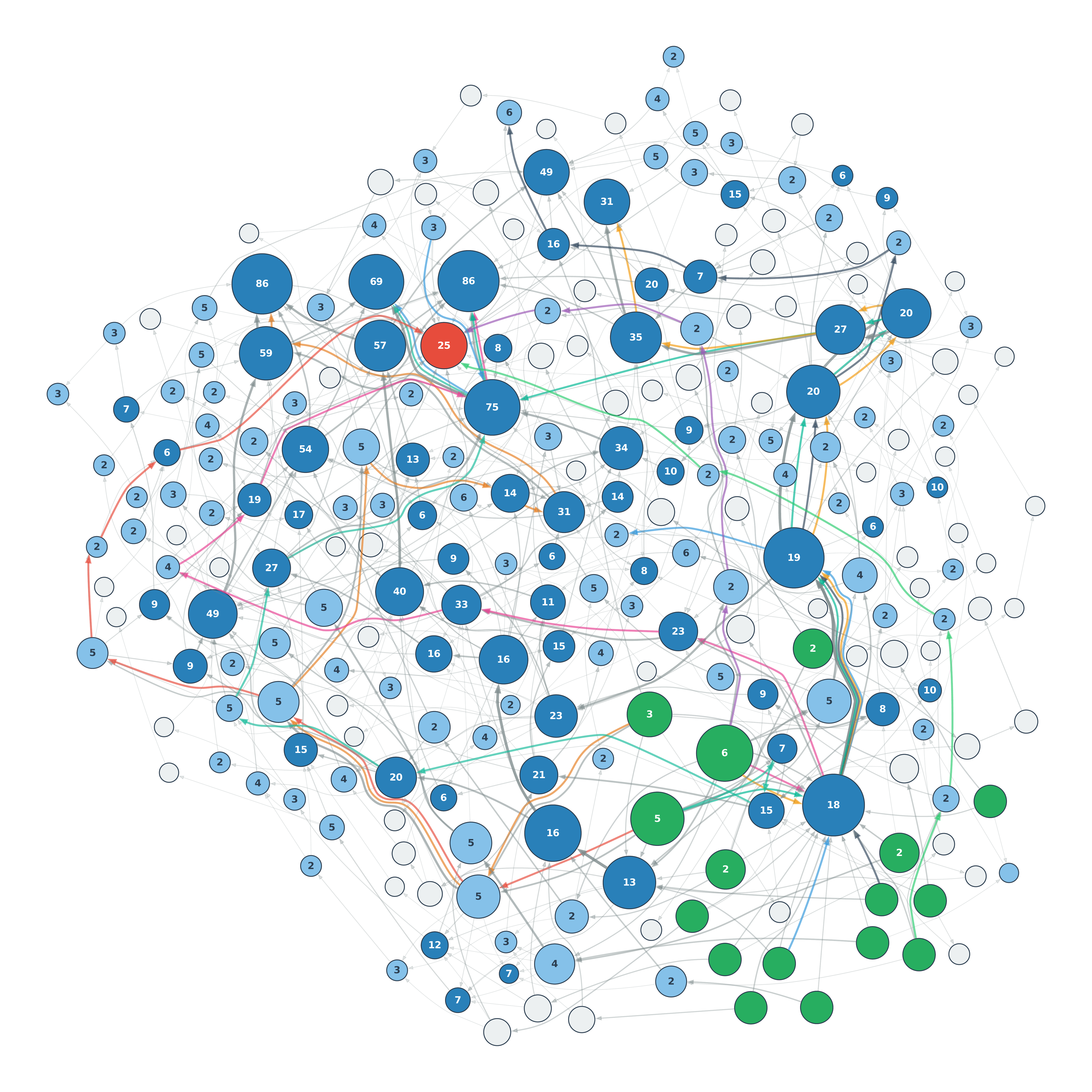}\hfill
\includegraphics[width=0.48\textwidth]{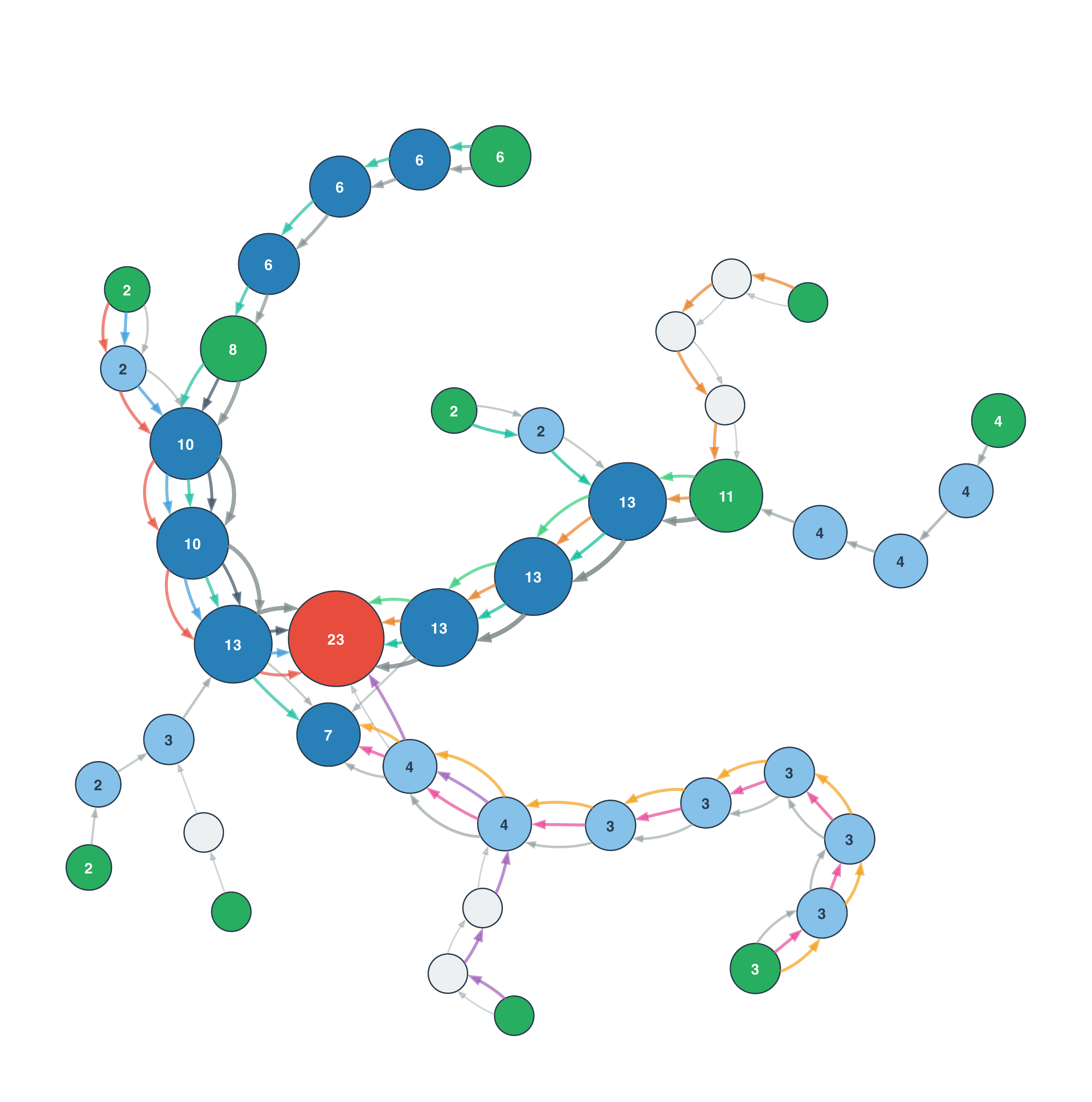}
\caption{}
\end{subfigure}
\caption{\textbf{Direct measurement of PSV compression during training of a small Transformer.} (\textbf{a}) State-graph representation of the $2\times 3$ sliding puzzle: $N=120$ legal positions form a $K=3$ regular graph; shown is a shortest solution path of distance 4. (\textbf{b}) Time evolution of success rate (top) and compression ratio $R_c$ (bottom). $R_c$ traces a V-shape: an initial drop driven by policy diversification, followed by a rebound and saturation in a high-compression regime (65--91\%) that signals the freezing-in of the topological skeleton. (\textbf{c}) PSV effective state-transition graph at early (left, epoch 10) and late (right, epoch 1000) training. The number of states contracts by roughly an order of magnitude, revealing a sparse multi-input, single-output inverse-tree structure.}
\label{fig:compression}
\end{figure}

We emphasise that the argument above establishes the \emph{existence} of an effective Markov representation as a logical consequence of finite parametric capacity, but the actual partition of histories into PSV classes depends on the trained parameters and can only be constructed numerically, not derived analytically. Direct enumeration of equivalence classes in real LLMs (with $10^9$--$10^{11}$ parameters) is currently infeasible. We therefore turn to a model system whose state space is exhaustively enumerable: the $2\times 3$ sliding puzzle ($N=120$ legal positions; the three actions $V$, $L$, $R$ are compound permutations that keep the blank in the centre column; Fig.~\ref{fig:compression}a), on which we train a minimal autoregressive Transformer with GRPO (experimental details in the supplementary material). The decisive advantage of this system is that the entire branching subtree below every decision point can be expanded explicitly, so the equivalence relation~(\ref{eq:psv}) can be checked directly and the full time evolution of the compression ratio can be tracked.

In this experiment, a prompt $Q$ is a start state and a history $h$ is a sequence of actions $(V,L,R)$ output so far. Each prefix $(Q,h)$ constitutes a decision point at which the model must choose the next action. Most of the ${\sim}120\times 3^{12}$ possible decision points are never visited during training because the policy assigns them negligible probability; including them would inflate $R_c$ with spurious classes. We therefore restrict the count to $\mathcal{S}_{\text{active}}$, the set of decision points whose ancestral chain of action probabilities is $\geq 0.05$, and define the compression ratio $R_c = 1 - N_c/|\mathcal{S}_{\text{active}}|$, where $N_c$ is the number of PSV equivalence classes on $\mathcal{S}_{\text{active}}$. $R_c\to 1$ means that distinct histories are highly equivalent in their future behaviour and the effective state space is heavily compressed.

The model is small enough (${\sim}14$K parameters) that GRPO converges reliably from a random initialization, without pretraining. This keeps the training dynamics purely RLVR-driven, free of any confound introduced by a pretraining phase.

The temporal evolution of $R_c$ (Fig.~\ref{fig:compression}b) traces a non-monotonic V-shape. Before training, the randomly initialized policy is near-uniform at every decision point---all nodes share essentially the same future, almost all states fall into a single PSV class, and $R_c$ is close to 1. \emph{Stage 1} (the rapid-learning window) is when the model begins to assign distinct action preferences to distinct states; equivalence classes split apart and $R_c$ plunges---this is the inevitable price of forming meaningful structure. In \emph{Stage 2}, $R_c$ rebounds and stabilises in a high-compression regime: reasoning paths of different problems start sharing intermediate nodes, and many \emph{distinct} histories collapse back onto a small number of classes---unlike the trivial initial compression, this one reflects genuine policy convergence. Throughout Stage 2, $R_c$ barely changes while the success rate continues to climb slowly. Visual inspection of the PSV graph (Fig.~\ref{fig:compression}c) suggests that the topological skeleton is largely laid down by the end of Stage~1, with later gains arising from refinement within that already-established structure rather than from further coarse-scale reorganisation.

As a scalar, $R_c$ only quantifies the degree of compression and says nothing about the topology of the compressed state space. Building the effective state-transition graph---PSV equivalence classes as nodes and high-probability transitions as directed edges---makes the post-training topology directly visible (Fig.~\ref{fig:compression}c): the dense, disordered initial graph contracts into a sparse \emph{multi-input, few-output} directed structure---many starting paths converge into a small number of shared hub nodes and then flow, along almost unique channels, to the target.

A widely accepted view in LLM research holds that ``compression is intelligence''~\citep{deletang2023language,huang2024compression}---the generalization capability of a model arises fundamentally from effective compression of the training data. The state-space compression observed in our exhaustively enumerable system is therefore very likely also at work during the RLVR training of LLMs built on the same Transformer architecture. Since the state space of a real LLM cannot be directly enumerated, the next sections develop the question from both theoretical and experimental sides: how is this sparse topology sculpted by the training dynamics of RLVR, and can its formation account for the macroscopic phenomena observed in real LLM training---two-stage dynamics, growth of reasoning chains, and catastrophic forgetting?

\section{The Concept Network: A Physical Picture of Slow Thinking}\label{sec:conet}

\begin{figure}[!htbp]
\centering
\begin{subfigure}[b]{0.32\textwidth}
\includegraphics[width=\textwidth]{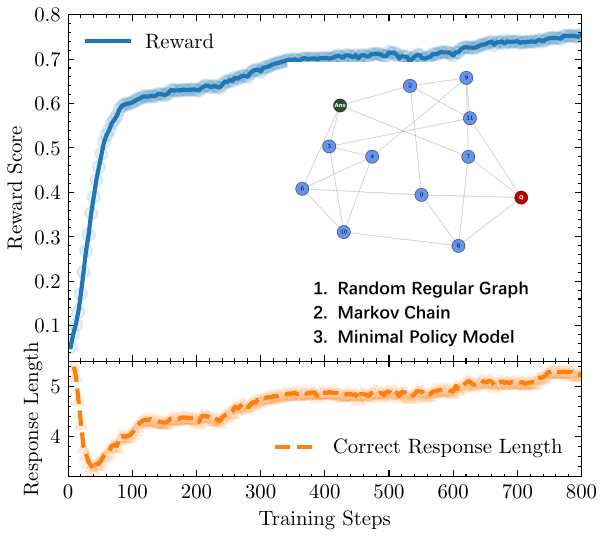}
\caption{CoNet}
\end{subfigure}
\hfill
\begin{subfigure}[b]{0.32\textwidth}
\includegraphics[width=\textwidth]{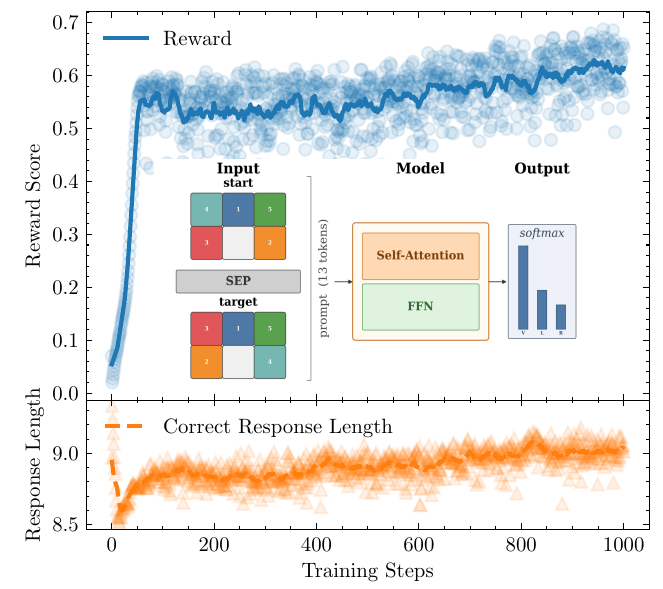}
\caption{Minimal Transformer}
\end{subfigure}
\hfill
\begin{subfigure}[b]{0.32\textwidth}
\includegraphics[width=\textwidth]{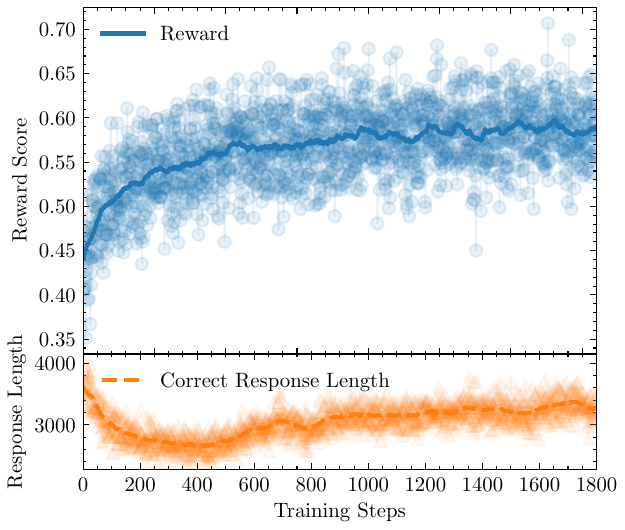}
\caption{DeepScaleR-1.5B}
\end{subfigure}
\caption{\textbf{The same two-stage dynamics span three scales of model.} (\textbf{a}) The minimal CoNet implementation: no neural network, transition probabilities as trainable parameters. (\textbf{b}) A 14K-parameter single-layer Transformer trained on the sliding puzzle. (\textbf{c}) DeepScaleR-1.5B, a 1.5-billion-parameter LLM. Top: reward; bottom: path/token length of correct responses. All three exhibit the universal Stage~1 (fast learning, shortening length) $\to$ Stage~2 (slow optimization, rebound in length) trajectory.}
\label{fig:training_dynamics}
\end{figure}

The PSV analysis of the previous section shows that, after RLVR training, the LLM admits an effective Markov representation. Within this representation the reasoning process is naturally a random walk on a sparse directed graph: each PSV equivalence class is a node, transition probabilities are determined implicitly by the model parameters, a single transition is one local computation (fast thinking), and a multi-step path from a question node to an answer node constitutes the full reasoning chain (slow thinking). This is the core of the Concept Network (CoNet) picture~\citep{cai2025learning}: slow thinking is fast thinking unfolded over time on a sparse network. This provides a different representation for RLVR from the reasoning-graph analyses discussed in~\citep{wang2024understandingreasoningabilitylanguage,minegishi2025topology}.

In a real Transformer, every transition probability is jointly determined by self-attention, multilayer perceptrons and other intricate components; at the level of the effective Markov representation, however, all such microscopic detail is compressed into a single transition probability $p(s'\,|\,s)$ between nodes. After pretraining, before any task-specific optimization, a great many outgoing edges at every node carry non-zero transition probability and the effective state-transition graph is a dense, high-entropy network---a random walker faces too many possible branches at every node to find a reliable path to the answer. RLVR's role is precisely to prune these transition probabilities systematically: through GRPO and other policy-gradient methods, edges along successful paths are reinforced and the rest are suppressed; the outgoing-edge entropy of every node decreases, so that the network converges from a high-entropy uniform distribution to a low-entropy structure dominated by a few edges.

The minimal implementation of the CoNet picture treats the transition probabilities themselves as trainable parameters. We start from a $K$-regular random graph $G(N,K)$ ($N$ nodes, each emitting exactly $K$ directed edges); each directed edge $s\to s'$ carries a trainable probability $p(s'\,|\,s)$ with $\sum_{s'}p(s'\,|\,s)=1$; reasoning is a random walk that starts at the question node $Q$ and samples transitions according to $p(s'\,|\,s)$, receiving reward 1 if it reaches the answer node $A$ and 0 otherwise; transition probabilities are updated by GRPO. This minimal implementation strips away every Transformer component---self-attention, MLPs, embeddings, positional encoding---and retains only the three essential ingredients of RLVR: path search on a sparse graph, shared transition probabilities, and a sparse reward at the end of the sequence.

The value of this minimal implementation is full observability: every transition probability at every node can be logged in full throughout training, and the subgraph of high-probability edges can be visualised directly. If the CoNet picture captures the core mechanism of RLVR, then the macroscopic dynamical signatures of the minimal implementation should also appear in any other system that satisfies the same conditions (sparse graph, shared parameters, sparse reward). The curve shapes depend on the initial policy and the number of training problems, and under different conditions---a pretrained model, a small problem set, or a single QA pair---they can look quite different~\citep{power2022grokkinggeneralizationoverfittingsmall,cai2025learning}. What persists across all these regimes is cross-model consistency: the three systems track each other regardless. Figure~\ref{fig:training_dynamics} shows a representative case---a high-entropy initial policy with many training problems---for CoNet (no neural network), a 14K-parameter Transformer on the sliding puzzle, and DSR-1.5B trained under the DeepScaleR protocol~\citep{deepscaler2025} using the verl framework~\citep{sheng2025hybridflow,verl_github}, covering five orders of magnitude in parameter count. All three display two-stage dynamics with shortening-then-lengthening responses. Among these shared features, the persistent growth of response length during Stage~2 is the most distinctive: unlike the two-stage reward profile, which is common to many optimization processes, the steady lengthening of reasoning chains is a geometric consequence of the inverse-tree picture (see below) and is not expected from generic convergence.

\section{Two Core Mechanisms: Merging and Frustration}\label{sec:mechanisms}

\begin{figure}[!htbp]
\centering
\begin{minipage}[b]{0.292\textwidth}
\centering
\begin{subfigure}[b]{\textwidth}
\includegraphics[width=\textwidth]{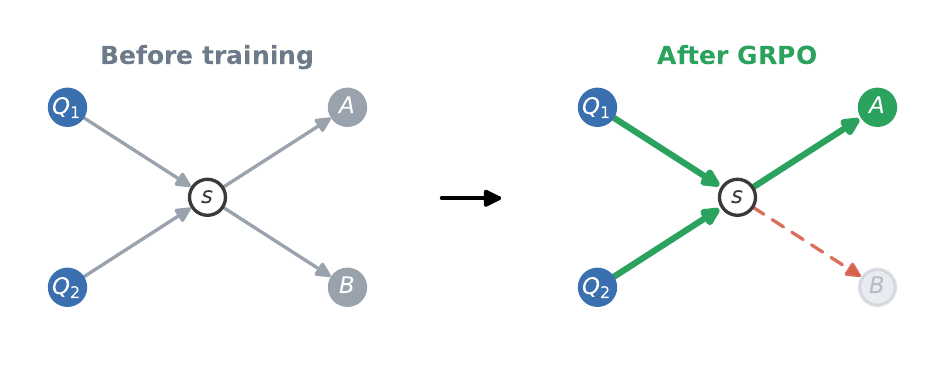}
\caption{Merging}
\label{fig:mechanisms_merge}
\end{subfigure}\\[0.4em]
\begin{subfigure}[b]{\textwidth}
\includegraphics[width=\textwidth]{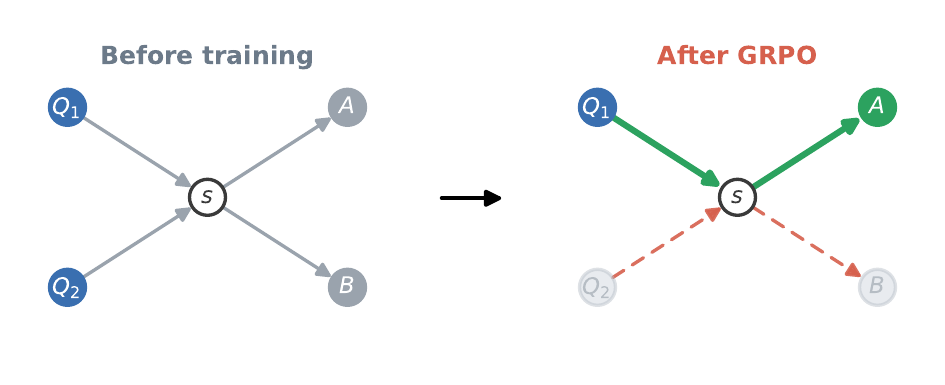}
\caption{Frustration}
\label{fig:mechanisms_frustration}
\end{subfigure}
\end{minipage}%
\hspace*{0.060\textwidth}%
\begin{subfigure}[b]{0.243\textwidth}
\includegraphics[width=\textwidth]{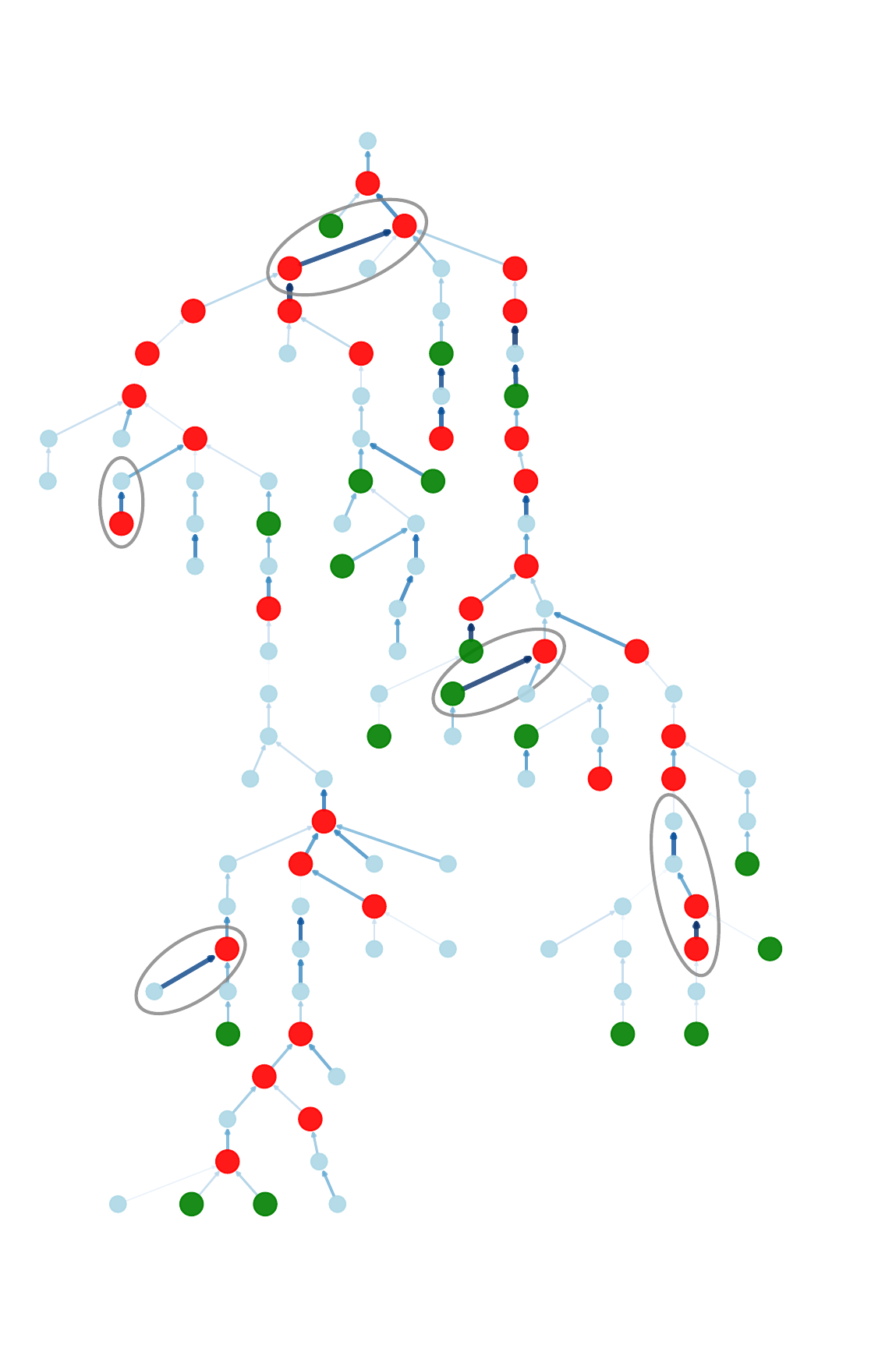}
\caption{Step 20}
\end{subfigure}%
\hspace*{0.012\textwidth}%
\begin{subfigure}[b]{0.243\textwidth}
\includegraphics[width=\textwidth]{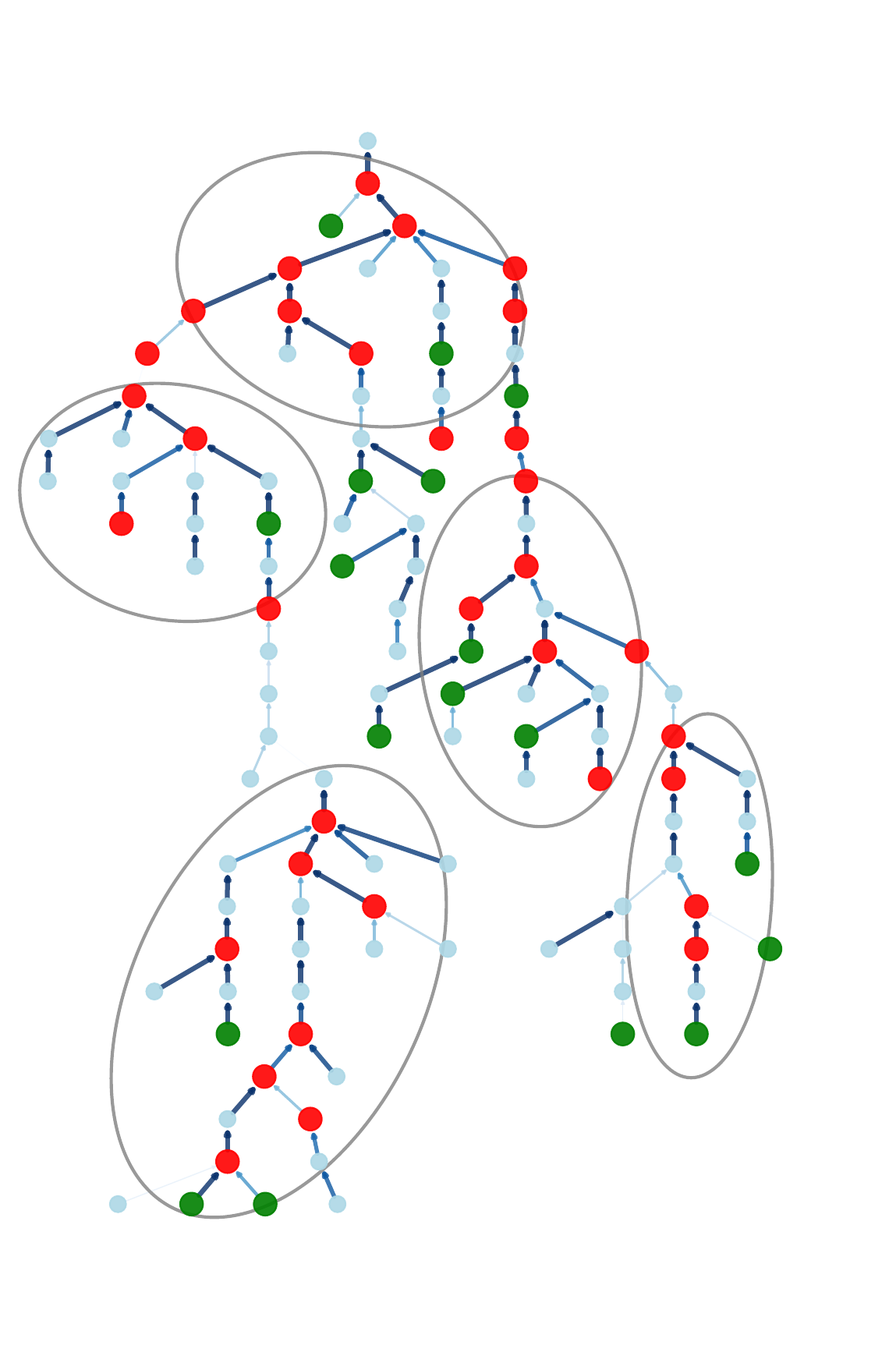}
\caption{Step 50}
\end{subfigure}%
\hspace*{0.012\textwidth}%
\begin{subfigure}[b]{0.365\textwidth}
\includegraphics[width=\textwidth]{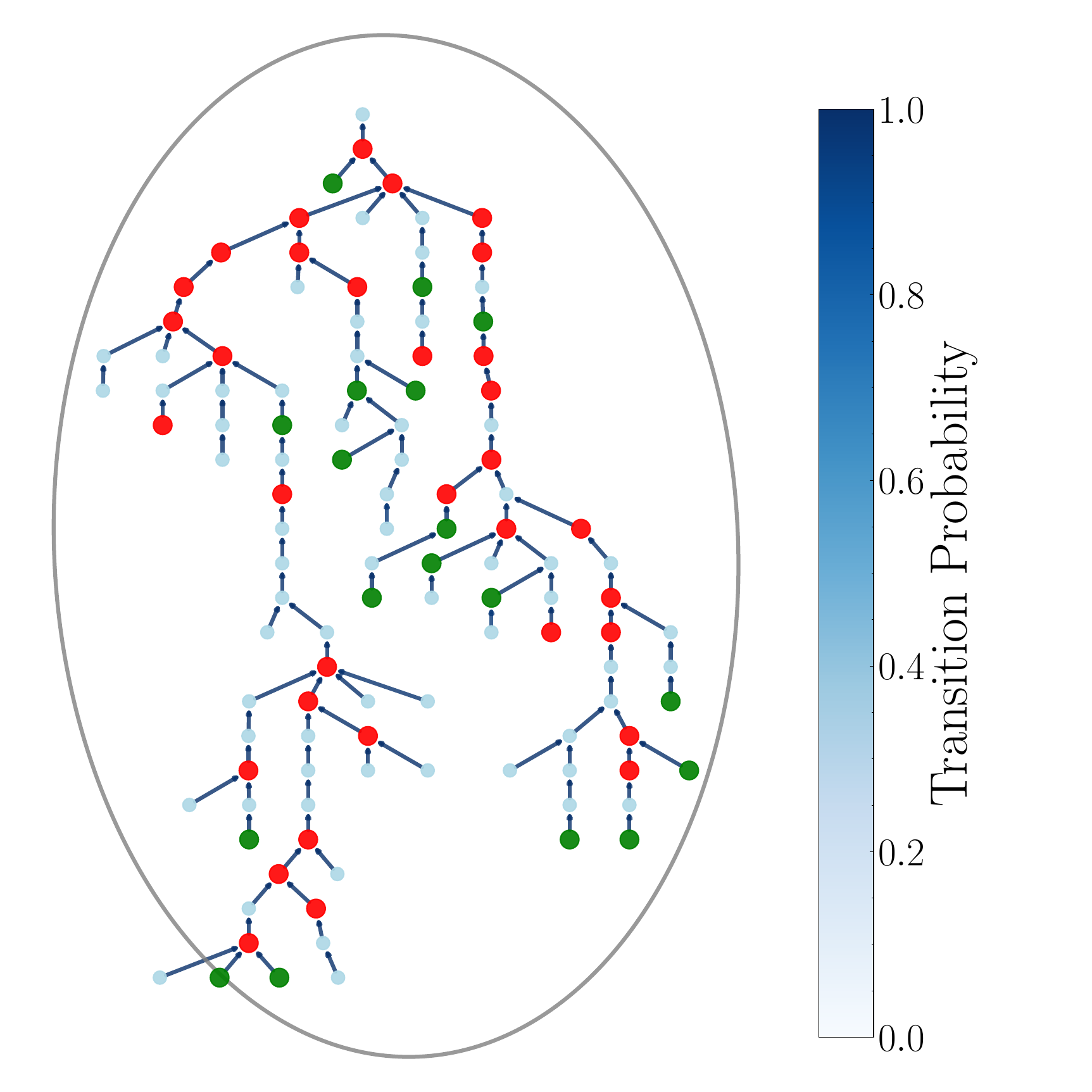}
\caption{Step 800}
\end{subfigure}
\caption{\textbf{The two microscopic mechanisms and the resulting inverse-tree freezing in CoNet.} (\textbf{a}) \emph{Merging}---two problems $Q_1$ and $Q_2$ pass through a shared node $s$ towards a common answer direction; GRPO reinforces the common outgoing edge and the paths weld together at $s$. (\textbf{b}) \emph{Frustration}---$Q_1$ requires $s\to A$ while $Q_2$ requires $s\to B$; positive feedback amplifies the dominant party and suppresses the alternative, systematically pruning the subordinate task's path (``learn-then-forget''). (\textbf{c}--\textbf{e}) Direct observation in CoNet: the largest inverse tree extracted at Step~800 is used as the substrate; earlier training steps are traced back on the same fixed layout. Red = answer nodes, green = question nodes; coloured intermediate nodes/edges indicate distinct weakly connected components, while grey semi-transparent nodes mark positions not yet active at the current step. (\textbf{c}) Step~20: multiple independent subtrees nucleate simultaneously. (\textbf{d}) Step~50: a dominant component takes over while a few small components have yet to merge. (\textbf{e}) Step~800: all nodes and edges coalesce into a single connected inverse tree with average degree ${\approx}2$.}
\label{fig:freezing}
\end{figure}

\begin{figure}[!htbp]
\centering
\begin{subfigure}[b]{0.48\textwidth}
\includegraphics[width=\textwidth]{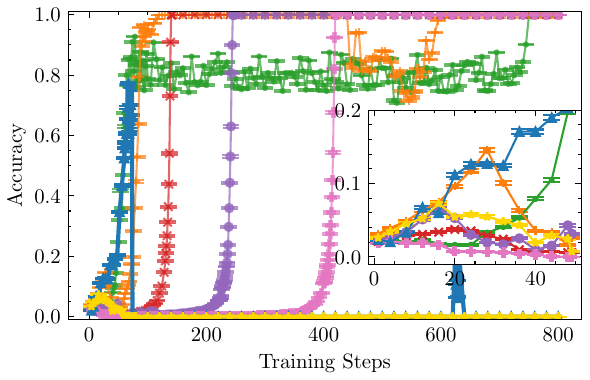}
\caption{CoNet}
\end{subfigure}
\hfill
\begin{subfigure}[b]{0.48\textwidth}
\includegraphics[width=\textwidth]{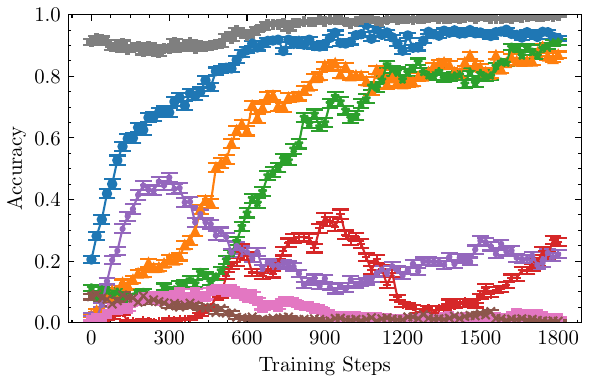}
\caption{LLM}
\end{subfigure}
\caption{\textbf{The ``learn-then-forget'' signature of frustration-induced forgetting appears jointly in CoNet and the LLM.} (\textbf{a}) Per-problem accuracy trajectories in CoNet; combined with the internal record of shared-node/edge conflicts, the non-monotonic ``learn-then-forget'' behaviour can be attributed directly to inter-task competition at shared nodes. (\textbf{b}) The same class of trajectories in DSR-1.5B, showing consistent non-monotonic patterns.}
\label{fig:frustration}
\end{figure}

The CoNet picture provides a mechanistic account of the PSV-graph contraction observed in Fig.~\ref{fig:compression}c: RLVR prunes a dense, high-entropy initial network into a sparse, low-entropy structure. Because the initial network is dense (Fig.~\ref{fig:compression}c, left), reasoning paths from different problems inevitably collide on shared nodes. Since multi-task RLVR forces all question--answer pairs to share a single set of transition probabilities $\{p(s'\,|\,s)\}$, each shared node $s$ receives competing demands on its outgoing edges from different tasks. The central question is how these shared-node transition probabilities evolve under RLVR (Fig.~\ref{fig:freezing}a,b).


A collision admits only two outcomes. If the two tasks need the \emph{same} outgoing edge from a shared node $s$---their constraints on $p(s'\,|\,s)$ are \emph{compatible}---then GRPO's positive feedback drives $p(s'\,|\,s)\to 1$ and the two paths fuse at $s$. The cumulative effect of merging is to forge multiple independent paths into a shared chain-like structure: many entry points reach the same shared node along different prefixes and then flow towards the answer along a single high-probability channel.

If the two tasks instead need \emph{different} outgoing edges from $s$---$Q_1$'s correct path requires $s\to A$ while $Q_2$ requires $s\to B$---their constraints are \emph{incompatible}. This is exactly the textbook setting of \emph{frustration} in statistical mechanics~\citep{toulouse1987theory}, in which, for instance, three spins on a triangular antiferromagnetic lattice cannot simultaneously satisfy all pairwise antiparallel requirements and the system is forced to compromise among competing constraints. The situation here is closely analogous: the $K$ outgoing edges at node $s$ face mutually exclusive demands from different tasks. GRPO's positive feedback magnifies any small initial advantage, the edge favoured by the dominant task is steadily reinforced and the alternatives are suppressed, and of the $K$ outgoing edges only one tends to survive. The path already learned by the subordinate task is thereby pruned in a systematic manner---this is \textbf{frustration-induced forgetting}.

Merging and frustration act jointly and point to a particular topology: a sparse \emph{multi-input, single-output} directed tree---merging creates ``multi-input,'' frustration enforces ``single-output.'' We call this structure an \textbf{inverse tree}: multiple paths flow in and a single path flows out, the reverse of a conventional rooted tree. The argument is general: it requires only a sparse state space, shared transition probabilities across tasks, and positive feedback from policy gradient---no specific architecture or training algorithm.

The minimal CoNet implementation makes this picture directly observable. By tracking every transition probability throughout training and thresholding at $p>0.95$, one can watch the two microscopic mechanisms reshape the network in real time. The result (Fig.~\ref{fig:freezing}) confirms the predicted topology: out-degree close to 1, in-degree of merging nodes $\gg 1$, average degree $\langle k\rangle\approx 2$. Figure~\ref{fig:freezing} shows this process at three training steps, marked on the CoNet training curve in Fig.~\ref{fig:training_dynamics}a. The formation mirrors the freezing of a supercooled liquid~\citep{avrami1939kinetics,kelton2010nucleation}: disconnected subtrees nucleate independently (Step~20, Stage~1), then weld together at shared nodes as merging and frustration sculpt the network (Step~50), and the nucleation-to-merging crossover coincides with the Stage~1-to-2 inflection; by Step~800 the network has frozen into a single connected inverse tree.

This sparse geometry has a direct consequence for response length. With out-degree close to 1 at almost every node, the graph has average degree $\langle k\rangle\approx 2$ and lacks short-range shortcuts. As the inverse tree extends to cover more question--answer pairs, geodesic distances grow monotonically---each edge is one coarse-grained reasoning step, so longer geodesics mean longer reasoning chains. When training begins from a high-entropy initial policy, the length trajectory is V-shaped: the early shortening reflects local optimization pruning redundant steps~\citep{fatemi2025concisereasoningreinforcementlearning}, while the subsequent rebound is the geometric signature of the sparse backbone having formed. The length curves across all three scales (Fig.~\ref{fig:training_dynamics}, bottom row) are consistent with this prediction.

The frustration mechanism can be tested independently by tracking per-problem accuracy. In Stage~1, subtrees nucleate in isolation and do not interfere. Once they meet at shared nodes, competition prunes the subordinate task's path, producing a characteristic non-monotonic ``learn-then-forget'' trajectory. The minimal CoNet makes this mechanism directly visible: by jointly recording per-problem accuracy and the internal conflict record at each shared node, the ``learn-then-forget'' curves can be attributed to specific inter-task competitions (Fig.~\ref{fig:frustration}a). The same signature appears in DSR-1.5B (Fig.~\ref{fig:frustration}b), providing experimental support in a real LLM. The Stage-1-to-2 inflection thus marks the \textbf{maximally frustrated state}---the moment of fiercest competition, when the inverse tree is most plastic.

\begin{figure}[!htbp]
\centering
\begin{subfigure}[b]{0.48\textwidth}
\includegraphics[width=\textwidth]{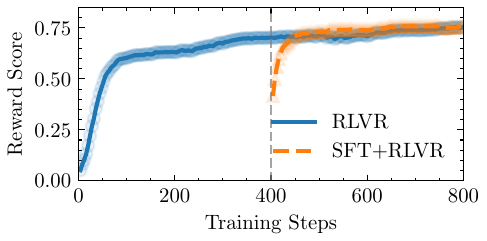}
\caption{CoNet}
\end{subfigure}
\hfill
\begin{subfigure}[b]{0.48\textwidth}
\includegraphics[width=\textwidth]{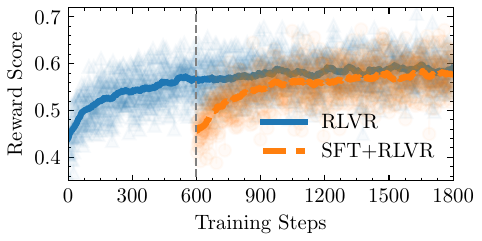}
\caption{DeepScaleR-1.5B}
\end{subfigure}

\vspace{0.3cm}
\begin{subfigure}[b]{0.3\textwidth}
\includegraphics[width=0.8\textwidth]{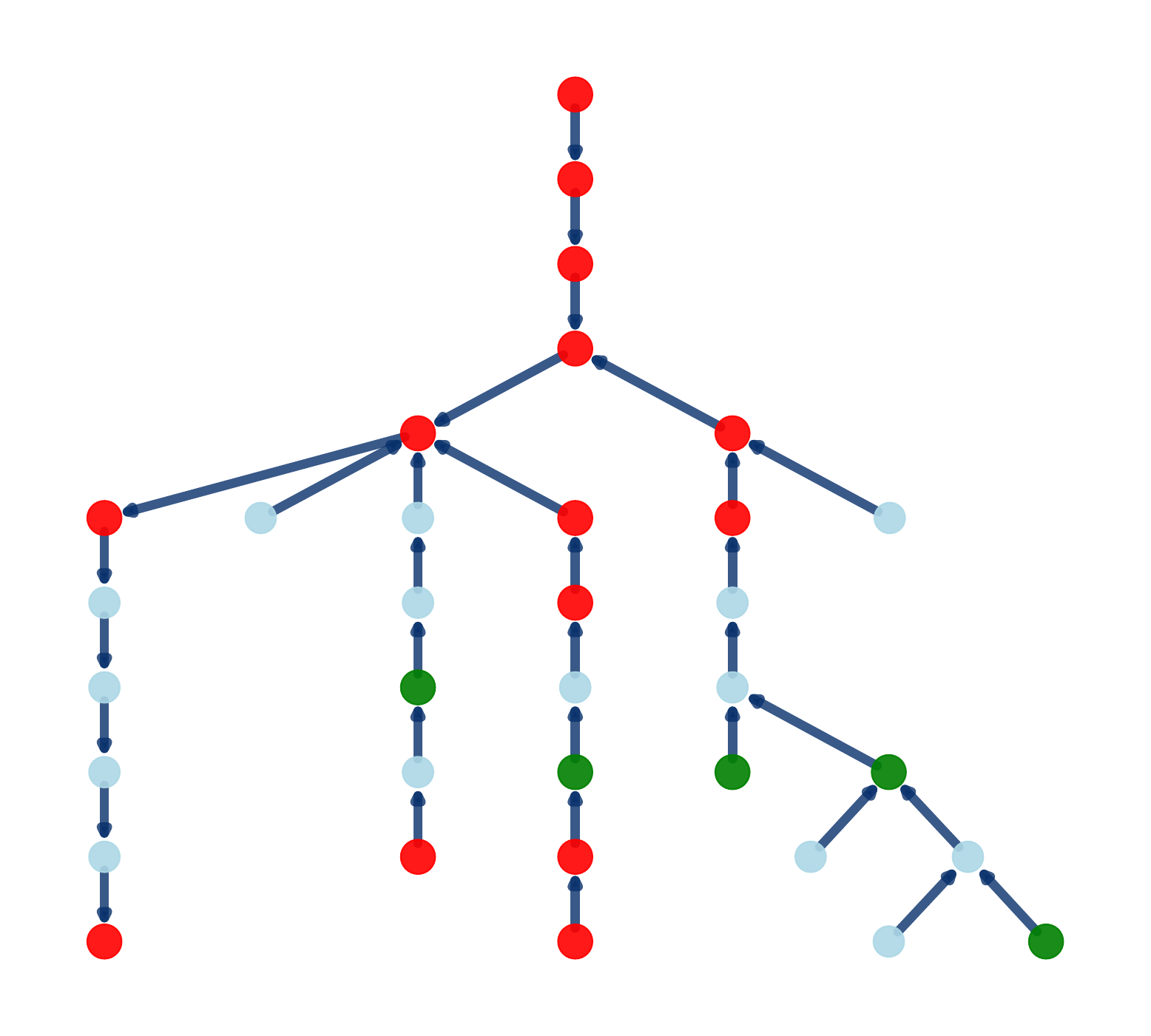}
\caption{Before SFT (Step 400)}
\end{subfigure}
\hfill
\begin{subfigure}[b]{0.3\textwidth}
\includegraphics[width=0.8\textwidth]{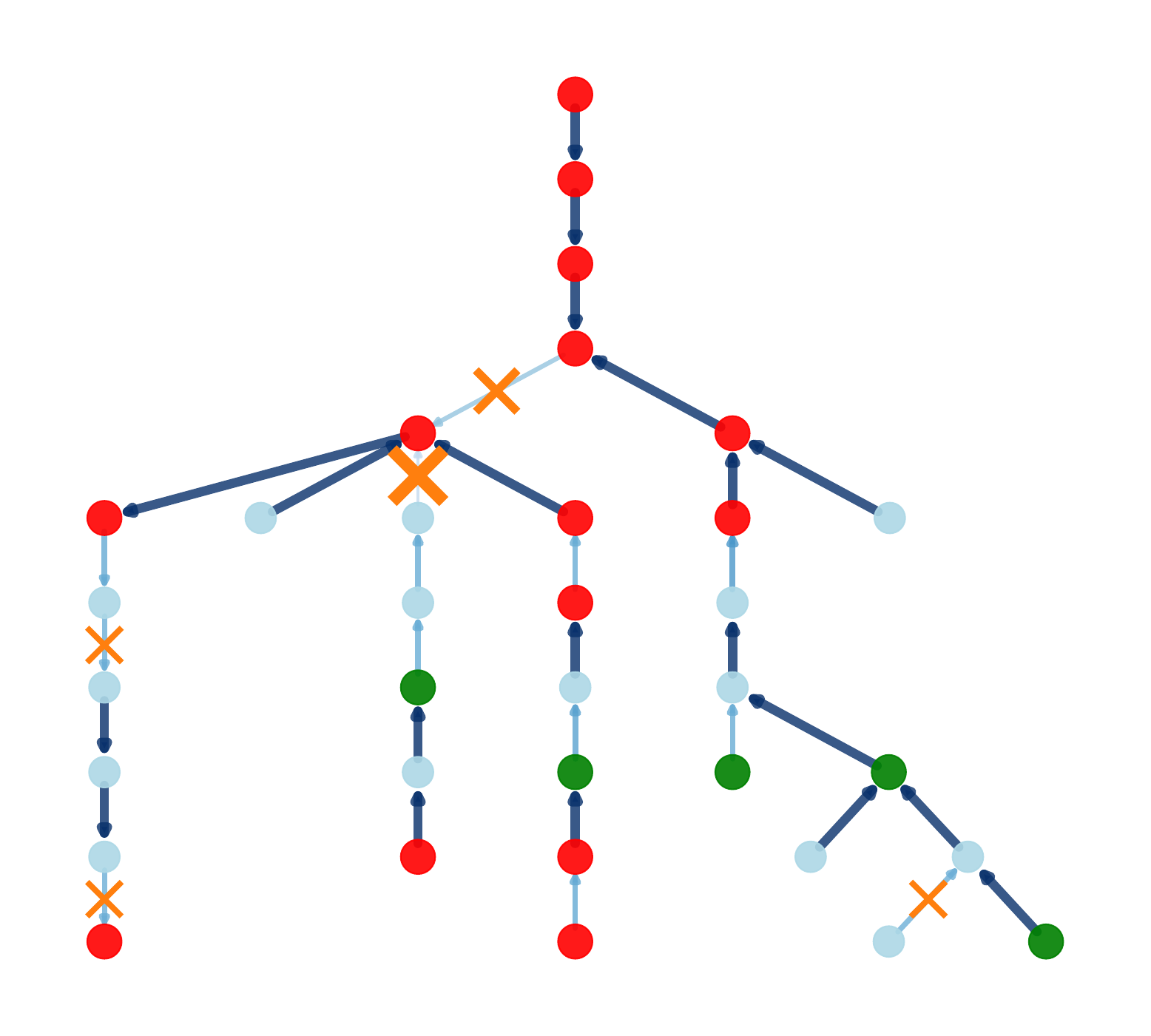}
\caption{After SFT (Step 400)}
\end{subfigure}
\hfill
\begin{subfigure}[b]{0.3\textwidth}
\includegraphics[width=0.8\textwidth]{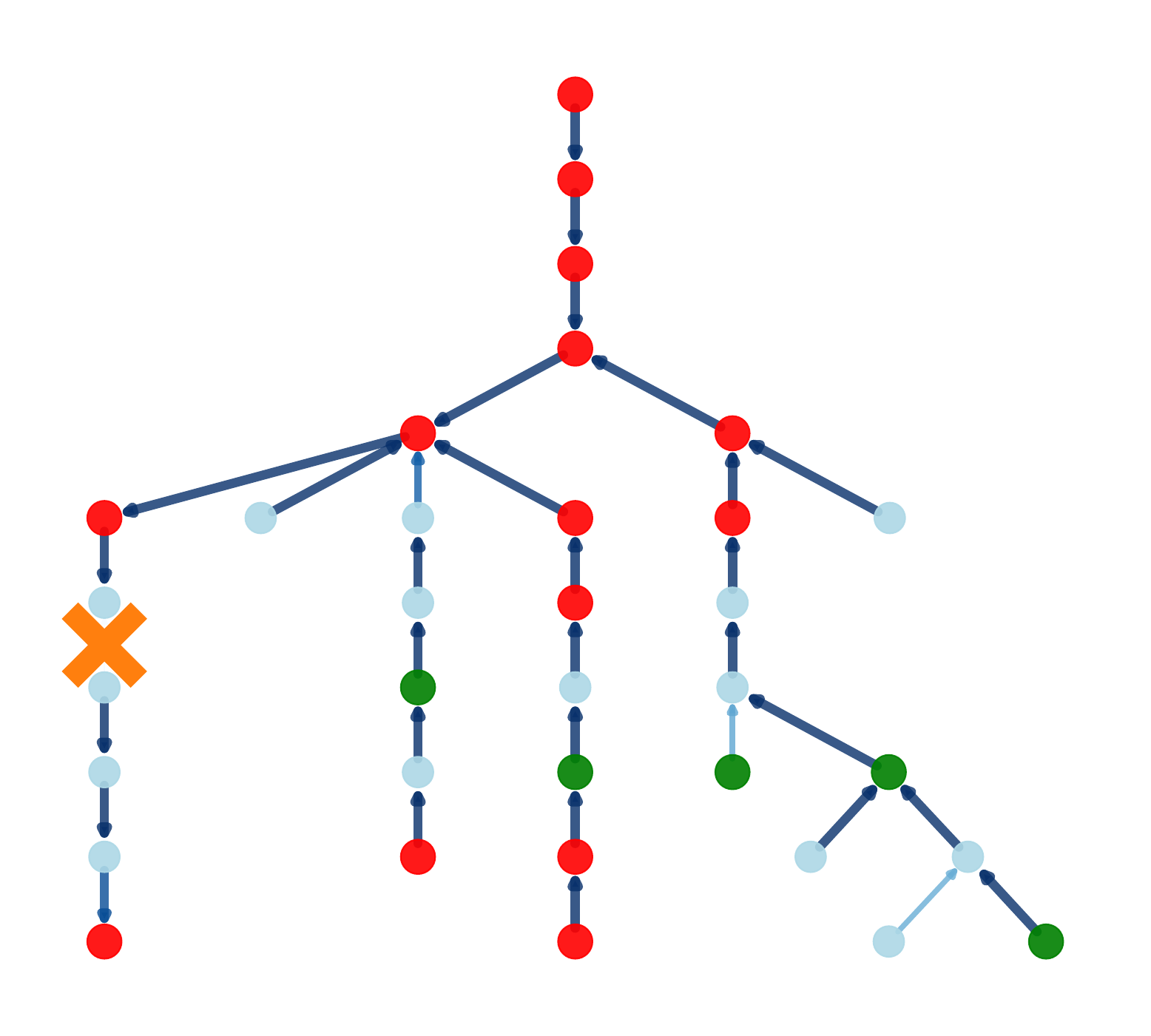}
\caption{Resumed RLVR (Step 450)}
\end{subfigure}
\caption{\textbf{SFT breaks but does not destroy: the signature response of a sparse-tree topology.} (\textbf{a, b}) Training curves of CoNet and DSR-1.5B: SFT applied during the slow-learning phase triggers an abrupt reward drop, and resumed RLVR leads to rapid recovery. (\textbf{c--e}) Microscopic evolution of the largest connected component in CoNet (edges with $p>0.95$; green = question, red = answer, blue = intermediate; orange crosses mark weakened transitions). SFT severs critical branching nodes (\textbf{d}), fragmenting the unified inverse tree (\textbf{c}$\to$\textbf{d}); resumed RLVR rapidly re-welds them (\textbf{e}).}
\label{fig:sft}
\end{figure}

\section{Catastrophic Forgetting and the Inverse Tree Picture}\label{sec:evidence}

A widely studied empirical puzzle in RLVR training is catastrophic forgetting~\citep{li2017learning,luo2025empiricalstudycatastrophicforgetting,ding2025improvedsupervisedfinetuninglarge}: inserting a supervised fine-tuning (SFT) intervention into the RLVR process causes a sharp drop in reasoning capability. Unlike RLVR, in which the model discovers solution paths through autonomous exploration, SFT requires the model to memorize prepared reasoning trajectories directly---maximizing token-by-token the conditional likelihood of the given trajectory, with no trial-and-error of the model's own.

The phenomenon has been reported in multiple independent studies~\citep{luo2025empiricalstudycatastrophicforgetting,ding2025improvedsupervisedfinetuninglarge,fernando2025understandingforgettingllmsupervised,matsutani2025rlsqueezessftexpands}. To amplify and study it under controlled conditions, we designed a dedicated SFT experiment (details in the supplementary material): the SFT trajectories are sourced from the model's own rollouts on hard problems in the training set, filtered by rejection sampling to retain only correct solutions---a deliberate data-preparation strategy intended to minimize the distributional gap between SFT data and the current policy. Applied during Stage 2 of DSR-1.5B's RLVR training (Fig.~\ref{fig:sft}b), this SFT triggers an immediate sharp reward drop; once SFT is stopped and RLVR resumes, performance rapidly returns to near its pre-SFT level. Even with SFT data drawn from the model's own correct trajectories---and the distributional gap therefore minimized---catastrophic forgetting cannot be avoided. This means that the root of forgetting lies not in any data distribution mismatch but in a conflict between the SFT mode of training itself and the internal structure that RLVR has already established.

The minimal CoNet implementation reproduces exactly the same response (Fig.~\ref{fig:sft}a) and additionally provides direct microscopic observation. Before SFT, the inverse tree is a single, fully connected structure (Fig.~\ref{fig:sft}c); after SFT, the tree fragments from the connected state into several disconnected subgraphs (Fig.~\ref{fig:sft}d); once RLVR resumes, the broken bridges are rapidly re-welded and the tree is restored (Fig.~\ref{fig:sft}e).

The inverse-tree picture supplies a unified explanation. Because almost every node has out-degree close to 1, the global connectivity of the inverse tree depends concentratedly on a small number of bridge nodes (high-in-degree merging points where multiple paths converge). SFT overwrites the transition probabilities only at the nodes its supervised trajectories visit, with no regard for the global tree structure already in place. Such trajectories are highly likely to traverse the key bridge nodes of the inverse tree, and an SFT overwrite at such a position effectively cuts the bridge---even when the overwrite happens to agree with the inverse-tree direction, the global modification of the outgoing-edge probabilities at that node disrupts other paths passing through the same node. The severity of forgetting is therefore set by the number of bridge nodes the SFT trajectory traverses, not by the quality of the SFT data itself---which explains why even using the model's own correct trajectories does not avert the destruction. The ``break-but-not-erase'' recovery, in turn, follows from the same topology: bridge nodes are few, hence fragile; but breakage does not destroy the subtrees themselves, so RLVR only needs to repair a small number of bridges to restore global connectivity.

\begin{figure}[!htbp]
\centering
\begin{subfigure}[b]{0.32\textwidth}
\includegraphics[width=\textwidth]{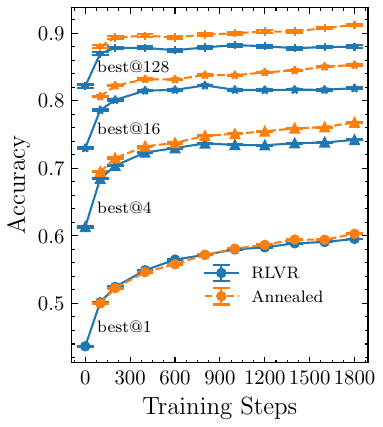}
\caption{In-distribution (512 problems)}
\end{subfigure}
\hfill
\begin{subfigure}[b]{0.32\textwidth}
\includegraphics[width=\textwidth]{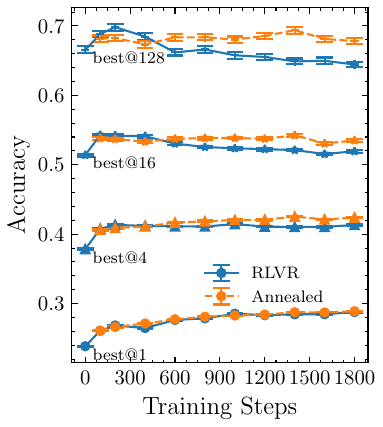}
\caption{OOD: Minerva}
\end{subfigure}
\hfill
\begin{subfigure}[b]{0.32\textwidth}
\includegraphics[width=\textwidth]{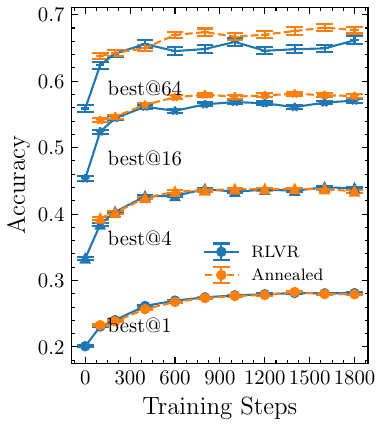}
\caption{OOD: AIME 2024/25}
\end{subfigure}
\caption{\textbf{Annealed-RLVR alleviates policy collapse, with the advantage concentrated at large-$k$ sampling.} (\textbf{a}) In-distribution 512 problems; (\textbf{b, c}) out-of-distribution Minerva and AIME 2024/25 benchmarks. Orange = Annealed-RLVR; blue = standard RLVR. The advantage of Annealed-RLVR grows monotonically with $k$, in agreement with the prediction that frustration-induced policy collapse is concentrated at large $k$.}
\label{fig:annealed}
\end{figure}

\section{Annealed-RLVR: From Physical Picture to Practical Method}\label{sec:annealed}

Policy collapse is one of the central practical challenges of RLVR training~\citep{yue2025doesreinforcementlearningreally,policy1,policy4}: as training proceeds, best@1 keeps improving while policy diversity is steadily lost---best@$k$, the probability that at least one of $k$ independent samples is correct, stagnates or even declines at large $k$ (blue curves in Fig.~\ref{fig:annealed}). Dominant paths grow ever stronger while the covered solution space contracts. Existing remedies rely largely on entropy regularization~\citep{cui2025entropymechanismreinforcementlearning,hao2025rethinkingentropyinterventionsrlvr,jiang2025rethinkingentropyregularizationlarge}---adding an entropy penalty to the loss to maintain policy stochasticity. Entropy regularization, however, is a \emph{global} and \emph{persistent} intervention: it cannot distinguish ``beneficial growth of certainty'' (the pruning of redundant paths in Stage 1) from ``harmful loss of diversity'' (the pruning of legitimate alternatives in Stage 2), and can only strike a compromise between the two.

The inverse-tree picture offers a different perspective: policy collapse is precisely the macroscopic accumulation of frustration-induced forgetting discussed above. The problem is not that the system's overall ``temperature'' is too low but that, throughout RLVR's monotonic cooling, the system has no opportunity to escape the configurations frozen in by frustration. Borrowing the idea of simulated annealing from condensed-matter physics~\citep{kirkpatrick1983optimization}---reheating during cooling to restore ergodicity---the inverse-tree picture suggests an intervention dimension that has rarely been considered in current methodology: \emph{structural timing}. The concrete scheme is to \textbf{apply a brief SFT ``heating'' at the moment of maximum frustration (the Stage-1-to-Stage-2 inflection) and then resume RLVR to let the system cool back down}. The timing is empirically located by the inflection of the training curve where the reward growth slows and the response length turns from shrinking to rebounding; the SFT data are the same self-rollout correct solutions used in the catastrophic-forgetting experiment (parameter values in the supplementary material). The central insight is not in SFT itself but in the fact that the same operation has opposite effects at different stages of the inverse tree's growth---a notion that can be generalised to other forms of policy perturbation.

Figure~\ref{fig:annealed} shows the experimental results on DSR-1.5B. The three panels report best@$k$ on, respectively, the in-distribution 512 training problems (a)~\citep{minerva1}, the out-of-distribution Minerva benchmark (b)~\citep{lewkowycz2022solving}, and the out-of-distribution AIME 2024/25 benchmark (c); the horizontal axis is the sample count $k$ and the vertical axis is best@$k$. On all three benchmarks, Annealed-RLVR (orange) outperforms standard RLVR (blue), with a striking pattern: the gap is small at $k=1$ but grows monotonically with $k$, peaking at $k=64$. This pattern carries a clear physical meaning: annealing restores the \emph{subordinate paths} that frustration had pruned; the dominant path (which determines best@1) is itself unaffected, so the improvement is concentrated along the policy-diversity axis. An improvement of best@$k$ at large $k$ means the model has regained multiple independent solution strategies rather than producing all samples along a single path. Notably, the improvement holds and is even more pronounced on the out-of-distribution benchmarks (Fig.~\ref{fig:annealed}b,c), indicating that the recovered diversity generalises and is not confined to the training set.

A causal test of timing further supports the picture. Postponing the same SFT to Step~400, by which time the inverse tree has fully solidified, not only fails to improve performance but in fact triggers catastrophic forgetting; resumed RLVR only returns the model to its pre-SFT level (Fig.~\ref{fig:sft}b). The same SFT data, the same amount of training---and yet the effect reverses depending solely on timing. This rules out the trivial interpretation that ``SFT supplied useful information'' and shows that the improvement of Annealed-RLVR is attributable specifically to recovering ergodicity in the policy space at the right structural moment.

\section{Conclusion and Outlook}\label{sec:discussion}

This work began from a physics question---what does RLVR change at the microscopic level so that long-range reasoning emerges in LLMs?---and arrived at a statistical-physics answer: RLVR drives the freezing-in of an inverse-tree structure on the model's effective state space. The formation process is dynamically akin to freezing, but the vast configurational degeneracy hints that the final state may sit closer to a glassy phase, awaiting further characterization.

The supporting logic of this answer can be read at three levels. The first is \emph{representation}: the self-referentiality of an autoregressive model and its finite parametric capacity force the effective state space to be compressed into a sparse Markov network; microscopic histories of order $V^t$ collapse into PSV equivalence classes, with compression ratios of 65--91\% measured in an exhaustively enumerable system. The second is \emph{dynamics}: on this sparse network, shared-parameter multi-task RLVR, through the merging of compatible paths and the frustrated competition of incompatible ones, freezes the dense initial network into a sparse, multi-input, single-output directed tree---an inverse tree---through three stages of nucleation $\to$ growth $\to$ merging. This dynamics displays qualitatively consistent macroscopic signatures across three systems whose parameter counts span five orders of magnitude. The third is \emph{prediction and intervention}: the geometric properties of the inverse tree directly yield three falsifiable predictions---reasoning chains lengthen with capability (a geodesic-distance constraint of the sparse tree), the break-and-recover response under SFT perturbation (topological fragility at bridge nodes), and frustration-driven policy collapse---all supported by experiment. Annealed-RLVR, designed on the strength of this picture, alleviates policy collapse while preserving exploration capacity, and a controlled experiment shows that the improvement is tied to the timing the mechanism specifies.

The significance of this picture extends beyond the explanation of specific empirical phenomena. It provides a methodology for \emph{translating} LLM reasoning into a statistical-physics problem: extracting effective degrees of freedom as PSV equivalence classes (analogous to coarse-graining), reducing reasoning to path search on a sparse graph (analogous to a Markov process), translating training dynamics into freezing on the graph (analogous to nucleation and growth), and recasting multi-task competition as frustration (analogous to spin glasses). Once this translation dictionary is in place, the existing toolbox of statistical physics---phase-transition theory, critical phenomena, and the replica method for disordered systems (the standard tool for handling vast quasi-degenerate states)---may all be brought to bear on the theoretical analysis of LLM reasoning.


\subsubsection*{Acknowledgements}
The authors thank Xianrui Ke, Daiqiu Mou, Yuansheng Cao, Weinan E, Haijun Zhou, Lei Wang, Hong Zhao and Ganqu Cui for inspiring discussions.


K.~C. and X.~C. are supported by the National Key Research and Development Program of China under Grant No.\ 2024YFA1408604 and the National Natural Science Foundation of China under Grants No.\ 12474245 and No.\ 12447103. P.~Z. is supported by Project 12325501 of the National Natural Science Foundation of China. Y.~D. and S.~H. are supported by the National Natural Science Foundation of China under Grants No.\ 12275263, the Quantum Science and Technology-National Science and Technology Major Project Grant No.\ 2021ZD0301900, and the Natural Science Foundation of Fujian Province of China under Grant No.\ 2023J02032. Z.~Y. is supported by the National Natural Science Foundation of China under Grants No.\ 12304288 and No.\ 12247101.

\bibliography{refs,refs_overflow}

\begin{thebibliography}{51}
\providecommand{\natexlab}[1]{#1}
\providecommand{\url}[1]{\texttt{#1}}
\expandafter\ifx\csname urlstyle\endcsname\relax
  \providecommand{\doi}[1]{doi: #1}\else
  \providecommand{\doi}{doi: \begingroup \urlstyle{rm}\Url}\fi

\bibitem[Anderson(1972)]{anderson1972more}
P.~W. Anderson.
\newblock More {{Is Different}}.
\newblock \emph{Science}, 177\penalty0 (4047):\penalty0 393--396, August 1972.
\newblock \doi{10.1126/science.177.4047.393}.

\bibitem[Avrami(1939)]{avrami1939kinetics}
Melvin Avrami.
\newblock Kinetics of phase change. {I.} general theory.
\newblock \emph{The Journal of chemical physics}, 7\penalty0 (12):\penalty0
  1103--1112, 1939.

\bibitem[Brown et~al.(2024)Brown, Juravsky, Ehrlich, Clark, Le, R{\'e}, and
  Mirhoseini]{brown2024large}
Bradley Brown, Jordan Juravsky, Ryan Ehrlich, Ronald Clark, Quoc~V Le,
  Christopher R{\'e}, and Azalia Mirhoseini.
\newblock Large language monkeys: scaling inference compute with repeated
  sampling.
\newblock arXiv:2407.21787, 2024.

\bibitem[Browne et~al.(2012)Browne, Powley, Whitehouse, Lucas, Cowling,
  Rohlfshagen, Tavener, Perez, Samothrakis, and Colton]{browne2012survey}
Cameron~B Browne, Edward Powley, Daniel Whitehouse, Simon~M Lucas, Peter~I
  Cowling, Philipp Rohlfshagen, Stephen Tavener, Diego Perez, Spyridon
  Samothrakis, and Simon Colton.
\newblock A survey of {Monte Carlo} tree search methods.
\newblock \emph{IEEE Transactions on Computational Intelligence and AI in
  games}, 4\penalty0 (1):\penalty0 1--43, 2012.

\bibitem[{ByteDance Seed Team and verl community}()]{verl_github}
{ByteDance Seed Team and verl community}.
\newblock verl: Volcano engine reinforcement learning for llms.
\newblock \url{https://github.com/volcengine/verl}.

\bibitem[Cai et~al.(2025)Cai, Hu, Wang, Huang, Zhang, Deng, and
  Chen]{cai2025learning}
Xiansheng Cai, Sihan Hu, Tao Wang, Yuan Huang, Pan Zhang, Youjin Deng, and Kun
  Chen.
\newblock Learning-at-criticality in large language models for quantum field
  theory and beyond.
\newblock \emph{Chinese Physics Letters}, 42\penalty0 (12):\penalty0 120002,
  2025.

\bibitem[Coulom(2006)]{coulom2006efficient}
R{\'e}mi Coulom.
\newblock Efficient selectivity and backup operators in {Monte-Carlo} tree
  search.
\newblock In \emph{International conference on computers and games}, pp.\
  72--83. Springer, 2006.

\bibitem[Cui et~al.(2025)Cui, Zhang, Chen, Yuan, Wang, Zuo, Li, Fan, Chen,
  Chen, Liu, Peng, Bai, Ouyang, Cheng, Zhou, and
  Ding]{cui2025entropymechanismreinforcementlearning}
Ganqu Cui, Yuchen Zhang, Jiacheng Chen, Lifan Yuan, Zhi Wang, Yuxin Zuo,
  Haozhan Li, Yuchen Fan, Huayu Chen, Weize Chen, Zhiyuan Liu, Hao Peng, Lei
  Bai, Wanli Ouyang, Yu~Cheng, Bowen Zhou, and Ning Ding.
\newblock The entropy mechanism of reinforcement learning for reasoning
  language models.
\newblock arXiv:2505.22617, 2025.
\newblock URL \url{https://arxiv.org/abs/2505.22617}.

\bibitem[{DeepSeek-AI}(2026)]{deepseekai2026deepseekv4}
{DeepSeek-AI}.
\newblock {DeepSeek-V4}: towards highly efficient million-token context
  intelligence.
\newblock
  \url{https://huggingface.co/deepseek-ai/DeepSeek-V4-Pro/blob/main/DeepSeek_V4.pdf},
  2026.
\newblock Tech report.

\bibitem[Del{\'e}tang et~al.(2023)Del{\'e}tang, Ruoss, Duquenne, Catt,
  Genewein, Mattern, Grau-Moya, Wenliang, Aitchison, Orseau,
  et~al.]{deletang2023language}
Gr{\'e}goire Del{\'e}tang, Anian Ruoss, Paul-Ambroise Duquenne, Elliot Catt,
  Tim Genewein, Christopher Mattern, Jordi Grau-Moya, Li~Kevin Wenliang,
  Matthew Aitchison, Laurent Orseau, et~al.
\newblock Language modeling is compression.
\newblock arXiv:2309.10668, 2023.

\bibitem[Ding \& Wang(2025)Ding and
  Wang]{ding2025improvedsupervisedfinetuninglarge}
Fei Ding and Baiqiao Wang.
\newblock Improved supervised fine-tuning for large language models to mitigate
  catastrophic forgetting.
\newblock arXiv:2506.09428, 2025.
\newblock URL \url{https://arxiv.org/abs/2506.09428}.

\bibitem[Dong et~al.(2025)Dong, Jiang, Tao, Liu, Zhang, Mou, Cao, Ma, Chen, Li,
  Jin, Huang, Li, and Li]{policy1}
Yihong Dong, Xue Jiang, Yongding Tao, Huanyu Liu, Kechi Zhang, Lili Mou, Rongyu
  Cao, Yingwei Ma, Jue Chen, Binhua Li, Zhi Jin, Fei Huang, Yongbin Li, and
  Ge~Li.
\newblock {{RL-PLUS}}: {{Countering Capability Boundary Collapse}} of {{LLMs}}
  in {{Reinforcement Learning}} with {{Hybrid-policy Optimization}}.
\newblock arXiv:2508.00222, July 2025.

\bibitem[Fatemi et~al.(2025)Fatemi, Rafiee, Tang, and
  Talamadupula]{fatemi2025concisereasoningreinforcementlearning}
Mehdi Fatemi, Banafsheh Rafiee, Mingjie Tang, and Kartik Talamadupula.
\newblock Concise reasoning via reinforcement learning.
\newblock arXiv:2504.05185, 2025.
\newblock URL \url{https://arxiv.org/abs/2504.05185}.

\bibitem[Fernando et~al.(2025)Fernando, Shen, Ram, Zhou, Samulowitz, Baracaldo,
  and Chen]{fernando2025understandingforgettingllmsupervised}
Heshan Fernando, Han Shen, Parikshit Ram, Yi~Zhou, Horst Samulowitz, Nathalie
  Baracaldo, and Tianyi Chen.
\newblock Understanding forgetting in llm supervised fine-tuning and preference
  learning -- a convex optimization perspective.
\newblock arXiv:2410.15483, 2025.
\newblock URL \url{https://arxiv.org/abs/2410.15483}.

\bibitem[Guo et~al.(2025)Guo, Yang, Zhang, Song, Wang, Zhu, Xu, Zhang, Ma, Bi,
  et~al.]{deepseek-r1}
Daya Guo, Dejian Yang, Haowei Zhang, Junxiao Song, Peiyi Wang, Qihao Zhu,
  Runxin Xu, Ruoyu Zhang, Shirong Ma, Xiao Bi, et~al.
\newblock {DeepSeek-R1} incentivizes reasoning in {LLMs} through reinforcement
  learning.
\newblock \emph{Nature}, 645\penalty0 (8081):\penalty0 633--638, 2025.

\bibitem[Hao et~al.(2025)Hao, Wang, Liu, Luo, Yu, Dong, Lin, Wang, and
  Chen]{hao2025rethinkingentropyinterventionsrlvr}
Zhezheng Hao, Hong Wang, Haoyang Liu, Jian Luo, Jiarui Yu, Hande Dong, Qiang
  Lin, Can Wang, and Jiawei Chen.
\newblock Rethinking entropy interventions in {RLVR}: an entropy change
  perspective.
\newblock arXiv:2510.10150, 2025.
\newblock URL \url{https://arxiv.org/abs/2510.10150}.

\bibitem[He et~al.(2025)He, Liu, Liu, Yan, Wang, Cheng, Zhang, Zhang, Xu, Shen,
  Li, Zeng, Wei, Cheng, An, Liu, and Zhou]{skywork-or1-2025}
Jujie He, Jiacai Liu, Chris~Yuhao Liu, Rui Yan, Chaojie Wang, Peng Cheng,
  Xiaoyu Zhang, Fuxiang Zhang, Jiacheng Xu, Wei Shen, Siyuan Li, Liang Zeng,
  Tianwen Wei, Cheng Cheng, Bo~An, Yang Liu, and Yahui Zhou.
\newblock Skywork open reasoner series.
\newblock
  \url{https://capricious-hydrogen-41c.notion.site/Skywork-Open-Reaonser-Series-1d0bc9ae823a80459b46c149e4f51680},
  2025.
\newblock Notion Blog.

\bibitem[Hendrycks et~al.(2021)Hendrycks, Burns, Kadavath, Arora, Basart, Tang,
  Song, and Steinhardt]{minerva1}
Dan Hendrycks, Collin Burns, Saurav Kadavath, Akul Arora, Steven Basart, Eric
  Tang, Dawn Song, and Jacob Steinhardt.
\newblock Measuring mathematical problem solving with the math dataset.
\newblock \emph{NeurIPS}, 2021.

\bibitem[Hua \& Zhang(2022)Hua and Zhang]{hua-zhang-2022-system}
Wenyue Hua and Yongfeng Zhang.
\newblock System 1 + system 2 = better world: Neural-symbolic chain of logic
  reasoning.
\newblock In Yoav Goldberg, Zornitsa Kozareva, and Yue Zhang (eds.),
  \emph{Findings of the Association for Computational Linguistics: EMNLP 2022},
  pp.\  601--612, Abu Dhabi, United Arab Emirates, December 2022. Association
  for Computational Linguistics.
\newblock \doi{10.18653/v1/2022.findings-emnlp.42}.
\newblock URL \url{https://aclanthology.org/2022.findings-emnlp.42/}.

\bibitem[Huang et~al.(2024)Huang, Zhang, Shan, and He]{huang2024compression}
Yuzhen Huang, Jinghan Zhang, Zifei Shan, and Junxian He.
\newblock Compression represents intelligence linearly.
\newblock arXiv:2404.09937, 2024.

\bibitem[Jaech et~al.(2024)Jaech, Kalai, Lerer, Richardson, El-Kishky, Low,
  Helyar, Madry, Beutel, Carney, et~al.]{jaech2024openai}
Aaron Jaech, Adam Kalai, Adam Lerer, Adam Richardson, Ahmed El-Kishky, Aiden
  Low, Alec Helyar, Aleksander Madry, Alex Beutel, Alex Carney, et~al.
\newblock {OpenAI} o1 system card.
\newblock arXiv:2412.16720, 2024.

\bibitem[Jiang et~al.(2025)Jiang, Li, Chen, Liu, Cheng, and
  Shao]{jiang2025rethinkingentropyregularizationlarge}
Yuxian Jiang, Yafu Li, Guanxu Chen, Dongrui Liu, Yu~Cheng, and Jing Shao.
\newblock Rethinking entropy regularization in large reasoning models.
\newblock arXiv:2509.25133, 2025.
\newblock URL \url{https://arxiv.org/abs/2509.25133}.

\bibitem[Kahneman(2011)]{kahneman2011thinking}
D.~Kahneman.
\newblock \emph{Thinking, Fast and Slow}.
\newblock Farrar, Straus and Giroux, New York, 2011.

\bibitem[Kelton \& Greer(2010)Kelton and Greer]{kelton2010nucleation}
Ken Kelton and Alan~Lindsay Greer.
\newblock \emph{Nucleation in condensed matter: applications in materials and
  biology}, volume~15.
\newblock Elsevier, 2010.

\bibitem[Kirkpatrick et~al.(1983)Kirkpatrick, Gelatt~Jr, and
  Vecchi]{kirkpatrick1983optimization}
Scott Kirkpatrick, C~Daniel Gelatt~Jr, and Mario~P Vecchi.
\newblock Optimization by simulated annealing.
\newblock \emph{Science}, 220\penalty0 (4598):\penalty0 671--680, 1983.

\bibitem[Lewkowycz et~al.(2022)Lewkowycz, Andreassen, Dohan, Dyer, Michalewski,
  Ramasesh, Slone, Anil, Schlag, Gutman-Solo, et~al.]{lewkowycz2022solving}
Aitor Lewkowycz, Anders Andreassen, David Dohan, Ethan Dyer, Henryk
  Michalewski, Vinay Ramasesh, Ambrose Slone, Cem Anil, Imanol Schlag, Theo
  Gutman-Solo, et~al.
\newblock Solving quantitative reasoning problems with language models.
\newblock \emph{Advances in neural information processing systems},
  35:\penalty0 3843--3857, 2022.

\bibitem[Li \& Hoiem(2017)Li and Hoiem]{li2017learning}
Zhizhong Li and Derek Hoiem.
\newblock Learning without forgetting.
\newblock \emph{IEEE transactions on pattern analysis and machine
  intelligence}, 40\penalty0 (12):\penalty0 2935--2947, 2017.

\bibitem[Littman et~al.(2001)Littman, Sutton, and Singh]{littman2001predictive}
Michael~L Littman, Richard~S Sutton, and Satinder Singh.
\newblock Predictive representations of state.
\newblock \emph{Advances in Neural Information Processing Systems},
  14:\penalty0 1555--1561, 2001.

\bibitem[Liu et~al.(2025)Liu, Chen, Li, Qi, Pang, Du, Lee, and Lin]{drgrpo}
Zichen Liu, Changyu Chen, Wenjun Li, Penghui Qi, Tianyu Pang, Chao Du, Wee~Sun
  Lee, and Min Lin.
\newblock Understanding {{R1-Zero-Like Training}}: {{A Critical Perspective}}.
\newblock arXiv:2503.20783, March 2025.

\bibitem[Luo et~al.(2025)Luo, Tan, Wong, Shi, Tang, Roongta, Cai, Luo, Li,
  Popa, and Stoica]{deepscaler2025}
Michael Luo, Sijun Tan, Justin Wong, Xiaoxiang Shi, William~Y. Tang, Manan
  Roongta, Colin Cai, Jeffrey Luo, Li~Erran Li, Raluca~Ada Popa, and Ion
  Stoica.
\newblock {DeepScaleR}: surpassing {O1}-{P}review with a 1.5{B} model by
  scaling {RL}.
\newblock
  \url{https://pretty-radio-b75.notion.site/DeepScaleR-Surpassing-O1-Preview-with-a-1-5B-Model-by-Scaling-RL-19681902c1468005bed8ca303013a4e2},
  2025.
\newblock Notion Blog.

\bibitem[Luo et~al.(2023)Luo, Yang, Meng, Li, Zhou, and
  Zhang]{luo2025empiricalstudycatastrophicforgetting}
Yun Luo, Zhen Yang, Fandong Meng, Yafu Li, Jie Zhou, and Yue Zhang.
\newblock An empirical study of catastrophic forgetting in large language
  models during continual fine-tuning.
\newblock arXiv:2308.08747, 2023.
\newblock URL \url{https://arxiv.org/abs/2308.08747}.

\bibitem[Matsutani et~al.(2025)Matsutani, Takashiro, Minegishi, Kojima,
  Iwasawa, and Matsuo]{matsutani2025rlsqueezessftexpands}
Kohsei Matsutani, Shota Takashiro, Gouki Minegishi, Takeshi Kojima, Yusuke
  Iwasawa, and Yutaka Matsuo.
\newblock Rl squeezes, sft expands: A comparative study of reasoning llms.
\newblock arXiv:2509.21128, 2025.
\newblock URL \url{https://arxiv.org/abs/2509.21128}.

\bibitem[Minegishi et~al.(2025)Minegishi, Furuta, Kojima, Iwasawa, and
  Matsuo]{minegishi2025topology}
Gouki Minegishi, Hiroki Furuta, Takeshi Kojima, Yusuke Iwasawa, and Yutaka
  Matsuo.
\newblock Topology of reasoning: Understanding large reasoning models through
  reasoning graph properties.
\newblock arXiv:2506.05744, 2025.

\bibitem[Power et~al.(2022)Power, Burda, Edwards, Babuschkin, and
  Misra]{power2022grokkinggeneralizationoverfittingsmall}
Alethea Power, Yuri Burda, Harri Edwards, Igor Babuschkin, and Vedant Misra.
\newblock Grokking: Generalization beyond overfitting on small algorithmic
  datasets.
\newblock arXiv:2201.02177, 2022.
\newblock URL \url{https://arxiv.org/abs/2201.02177}.

\bibitem[Shao et~al.(2024)Shao, Wang, Zhu, Xu, Song, Bi, Zhang, Zhang, Li, Wu,
  and Guo]{shao_deepseekmath_2024}
Zhihong Shao, Peiyi Wang, Qihao Zhu, Runxin Xu, Junxiao Song, Xiao Bi, Haowei
  Zhang, Mingchuan Zhang, Y.~K. Li, Y.~Wu, and Daya Guo.
\newblock {DeepSeekMath}: {Pushing} the {Limits} of {Mathematical} {Reasoning}
  in {Open} {Language} {Models}.
\newblock arXiv:2402.03300, April 2024.
\newblock URL \url{http://arxiv.org/abs/2402.03300}.
\newblock arXiv:2402.03300 [cs].

\bibitem[Sheng et~al.(2025)Sheng, Zhang, Ye, Wu, Zhang, Zhang, Peng, Lin, and
  Wu]{sheng2025hybridflow}
Guangming Sheng, Chi Zhang, Zilingfeng Ye, Xibin Wu, Wang Zhang, Ru~Zhang,
  Yanghua Peng, Haibin Lin, and Chuan Wu.
\newblock Hybridflow: A flexible and efficient rlhf framework.
\newblock In \emph{Proceedings of the Twentieth European Conference on Computer
  Systems}, pp.\  1279--1297, 2025.

\bibitem[Silver et~al.(2016)Silver, Huang, Maddison, Guez, Sifre, Van
  Den~Driessche, Schrittwieser, Antonoglou, Panneershelvam, Lanctot,
  et~al.]{silver2016mastering}
David Silver, Aja Huang, Chris~J Maddison, Arthur Guez, Laurent Sifre, George
  Van Den~Driessche, Julian Schrittwieser, Ioannis Antonoglou, Veda
  Panneershelvam, Marc Lanctot, et~al.
\newblock Mastering the game of {Go} with deep neural networks and tree search.
\newblock \emph{Nature}, 529\penalty0 (7587):\penalty0 484--489, 2016.

\bibitem[Singh et~al.(2025)Singh, Fry, Perelman, Tart, Ganesh, El-Kishky,
  McLaughlin, Low, Ostrow, Ananthram, et~al.]{singh2025openai}
Aaditya Singh, Adam Fry, Adam Perelman, Adam Tart, Adi Ganesh, Ahmed El-Kishky,
  Aidan McLaughlin, Aiden Low, AJ~Ostrow, Akhila Ananthram, et~al.
\newblock {OpenAI} {GPT-5} system card.
\newblock arXiv:2601.03267, 2025.

\bibitem[Snell et~al.(2024)Snell, Lee, Xu, and Kumar]{snell2024scaling}
Charlie Snell, Jaehoon Lee, Kelvin Xu, and Aviral Kumar.
\newblock Scaling {LLM} test-time compute optimally can be more effective than
  scaling model parameters.
\newblock arXiv:2408.03314, 2024.

\bibitem[Team et~al.(2025)Team, Du, Gao, Xing, Jiang,
  et~al.]{teamKimiK15Scaling2025}
Kimi Team, Angang Du, Bofei Gao, Bowei Xing, Changjiu Jiang, et~al.
\newblock Kimi k1.5: {{Scaling Reinforcement Learning}} with {{LLMs}}.
\newblock arXiv:2501.12599, March 2025.

\bibitem[Team et~al.(2026)Team, Bai, Bai, Bao, Cai, Cao, Charles, Che, Chen,
  Chen, et~al.]{team2026kimi}
Kimi Team, Tongtong Bai, Yifan Bai, Yiping Bao, SH~Cai, Yuan Cao, Y~Charles,
  HS~Che, Cheng Chen, Guanduo Chen, et~al.
\newblock {Kimi K2.5}: Visual agentic intelligence.
\newblock arXiv:2602.02276, 2026.

\bibitem[Toulouse et~al.(1987)]{toulouse1987theory}
G~Toulouse et~al.
\newblock Theory of the frustration effect in spin glasses: I.
\newblock \emph{Spin Glass Theory and Beyond: An Introduction to the Replica
  Method and Its Applications}, 9:\penalty0 99, 1987.

\bibitem[Vaswani et~al.(2017)Vaswani, Shazeer, Parmar, Uszkoreit, Jones, Gomez,
  Kaiser, and Polosukhin]{vaswani2017attention}
Ashish Vaswani, Noam Shazeer, Niki Parmar, Jakob Uszkoreit, Llion Jones,
  Aidan~N Gomez, {\L}ukasz Kaiser, and Illia Polosukhin.
\newblock Attention is all you need.
\newblock \emph{Advances in neural information processing systems}, 30, 2017.

\bibitem[Wang et~al.(2025)Wang, Yu, Gao, Zheng, Liu, Lu, Dang, Chen, Yang,
  Zhang, et~al.]{wang2025beyond}
Shenzhi Wang, Le~Yu, Chang Gao, Chujie Zheng, Shixuan Liu, Rui Lu, Kai Dang,
  Xionghui Chen, Jianxin Yang, Zhenru Zhang, et~al.
\newblock Beyond the 80/20 rule: High-entropy minority tokens drive effective
  reinforcement learning for {LLM} reasoning.
\newblock arXiv:2506.01939, 2025.

\bibitem[Wang et~al.(2024)Wang, Amayuelas, Zhang, Pan, Chen, and
  Wang]{wang2024understandingreasoningabilitylanguage}
Xinyi Wang, Alfonso Amayuelas, Kexun Zhang, Liangming Pan, Wenhu Chen, and
  William~Yang Wang.
\newblock Understanding reasoning ability of language models from the
  perspective of reasoning paths aggregation.
\newblock arXiv:2402.03268, 2024.
\newblock URL \url{https://arxiv.org/abs/2402.03268}.

\bibitem[Yang et~al.(2025)Yang, Li, Yang, Zhang, Hui, et~al.]{qwen3}
An~Yang, Anfeng Li, Baosong Yang, Beichen Zhang, Binyuan Hui, et~al.
\newblock Qwen3 technical report.
\newblock arXiv:2505.09388, 2025.

\bibitem[Yang et~al.(2026)Yang, Xu, Xiong, and E]{yang2026slowthinking}
Hongkang Yang, Zhiqin~John Xu, Feiyu Xiong, and Weinan E.
\newblock A first-principles theory of slow thinking and active perception.
\newblock Preprint, available at ResearchGate, 2026.

\bibitem[Yu et~al.(2025)Yu, Zhang, Zhu, Yuan, Zuo, et~al.]{yu_dapo_2025}
Qiying Yu, Zheng Zhang, Ruofei Zhu, Yufeng Yuan, Xiaochen Zuo, et~al.
\newblock {DAPO}: {An} {Open}-{Source} {LLM} {Reinforcement} {Learning}
  {System} at {Scale}.
\newblock arXiv:2503.14476, March 2025.
\newblock URL \url{http://arxiv.org/abs/2503.14476}.
\newblock arXiv:2503.14476 [cs] version: 1.

\bibitem[Yue et~al.(2025)Yue, Chen, Lu, Zhao, Wang, Yue, Song, and
  Huang]{yue2025doesreinforcementlearningreally}
Yang Yue, Zhiqi Chen, Rui Lu, Andrew Zhao, Zhaokai Wang, Yang Yue, Shiji Song,
  and Gao Huang.
\newblock Does reinforcement learning really incentivize reasoning capacity in
  llms beyond the base model?
\newblock arXiv:2504.13837, 2025.
\newblock URL \url{https://arxiv.org/abs/2504.13837}.

\bibitem[Zeng et~al.(2026)Zeng, Lv, Hou, Du, Zheng, Chen, Yin, Ge, Huang, Xie,
  et~al.]{zeng2026glm}
Aohan Zeng, Xin Lv, Zhenyu Hou, Zhengxiao Du, Qinkai Zheng, Bin Chen, Da~Yin,
  Chendi Ge, Chenghua Huang, Chengxing Xie, et~al.
\newblock {GLM-5}: from vibe coding to agentic engineering.
\newblock arXiv:2602.15763, 2026.

\bibitem[Zheng et~al.(2025)Zheng, Zhao, and Chen]{policy4}
Haizhong Zheng, Jiawei Zhao, and Beidi Chen.
\newblock Prosperity before collapse: How far can off-policy rl reach with
  stale data on llms?, 2025.
\newblock URL \url{https://arxiv.org/abs/2510.01161}.

\end{thebibliography}
\bibliographystyle{iclr2026_conference}

\appendix

\section{Sliding-Puzzle Experiment: Architecture, Training, and Compression Measurement}
\label{app:slidingpuzzle}

This section gives the full experimental specification for the sliding-puzzle Transformer used in §EFFECTIVE MARKOV REPRESENTATION of the main text, on which the PSV compression ratio $R_c$ is measured directly.

\subsection{Environment: $2\times 3$ Sliding Puzzle as a Reasoning Task}
\label{app:slidingpuzzle:env}

The $2\times 3$ board contains five labelled tiles ($1,\ldots,5$) and one blank. To make the dynamics a regular graph we restrict the blank to the middle column (positions $1$ and $4$ in row-major indexing) at all times. Enumerating the $5! = 120$ permutations of tiles with the blank at each of the two middle-column positions gives $240$ candidate configurations, of which exactly $120$ form a single connected component under the action set below (the other $120$ are unreachable due to permutation parity). The reachable set has $|\mathcal{S}|=N=120$ states.

The action set $\mathcal{A}=\{\mathrm{V},\mathrm{L},\mathrm{R}\}$ consists of three composite moves that all preserve the ``blank-in-middle-column'' invariant:
\begin{itemize}
\item $\mathrm{V}$ (\emph{vertical swap}): swap the blank between positions $1$ and $4$.
\item $\mathrm{L}$ (\emph{left rotation}): three-cycle the tiles in the left half of the board.
\item $\mathrm{R}$ (\emph{right rotation}): three-cycle the tiles in the right half of the board.
\end{itemize}
Every legal state admits all three actions, so the resulting state graph is $K=3$-regular (and bipartite under this generating set; cf.\ Fig.~1a of the main text).

A reasoning task is a question--answer pair $(Q,A)$ consisting of a starting state $Q$ and a target state $A$. For each fixed shortest-path distance $D\in\{4,6\}$ we draw $(Q,A)$ uniformly from all ordered state pairs satisfying $\mathrm{BFS}\text{-}\mathrm{dist}(Q,A)=D$; the first $B=500$ such pairs are used for training and the next $200$ for validation, so train and validation sets are disjoint subsets of the same candidate pool. The two reported settings (dist$=4$, dist$=6$; Fig.~\ref{fig:rc_rounddigits}) correspond to two independent training experiments, not a curriculum.

The model interacts with the environment by emitting a sequence of action tokens $a_1,a_2,\ldots$; the state advances deterministically and the episode terminates either when the answer state $A$ is reached (reward $1$) or the maximum length $L_{\max}=12$ is hit (reward $0$).

\subsection{Transformer Architecture (14K Parameters)}
\label{app:slidingpuzzle:arch}

The policy $\pi_\theta(a_{t+1}\mid Q,a_1,\ldots,a_t)$ is parameterised by a single-layer decoder-only Transformer with $\sim 14{,}000$ parameters in total. The configuration is summarised in Table~\ref{tab:transformer_arch}.

\begin{table}[h]
\centering
\caption{Architecture of the sliding-puzzle Transformer.}
\label{tab:transformer_arch}
\begin{tabular}{ll}
\toprule
Component & Value \\
\midrule
Number of decoder layers & 1 \\
Embedding / hidden dimension $d_{\text{model}}$ & 32 \\
Number of attention heads & 2 (head dim $d_h=16$) \\
Feed-forward dimension $d_{\text{ff}}$ & 128 \\
Activation & ReLU \\
Dropout & 0 \\
Position encoding & learned, \texttt{nn.Embedding}$(L_{\max}^{\text{seq}}, d_{\text{model}})$ \\
Token-type embedding & 2 types (state, action), additive \\
Maximum sequence length $L_{\max}^{\text{seq}}$ & 37 \\
Tile vocabulary & 7 (six tile ids $0,\ldots,5$ plus a $\textsc{sep}$ token) \\
Action vocabulary & 3 (independent of tile vocabulary) \\
Output head & Linear($d_{\text{model}}=32 \to 3$) \\
Total parameter count & $\sim 14{,}000$ \\
\bottomrule
\end{tabular}
\end{table}

The tile and action vocabularies use independent embedding tables; the model never tokenises an action with the tile vocabulary or vice versa. The prompt is encoded as
\begin{equation}
\textsc{prompt}\;=\;[\,Q_1,Q_2,\ldots,Q_6,\ \textsc{sep},\ A_1,A_2,\ldots,A_6\,] \qquad (\text{length}\ P=13),
\end{equation}
where $(Q_1,\ldots,Q_6)$ and $(A_1,\ldots,A_6)$ are the row-major tile sequences of the question and answer states. After the prompt, the model autoregressively appends action tokens (each of which contributes its action embedding plus the position embedding plus the action token-type embedding). A standard causal mask is applied throughout, and the next-action logits are read from the position of the most recently emitted action token.

\subsection{GRPO Training Protocol}
\label{app:slidingpuzzle:training}

The policy is trained by Group Relative Policy Optimization (GRPO) starting from a randomly initialised network --- no pretraining is used, since the entire reachable state space ($N=120$) is small enough that the policy can be learned from scratch. Hyperparameters are listed in Table~\ref{tab:transformer_training}.

\begin{table}[h]
\centering
\caption{Training hyperparameters for the sliding-puzzle Transformer.}
\label{tab:transformer_training}
\begin{tabular}{ll}
\toprule
Setting & Value \\
\midrule
Optimiser & AdamW \\
Learning rate & $3\times 10^{-4}$ \\
Weight decay & $0$ \\
PPO clip $\varepsilon_{\text{clip}}$ & $0.2$ \\
KL coefficient $\beta$ & $0$ (main $R_c$ experiments) \\
Inner update epochs $K$ per GRPO step & $4$ \\
Mini-batch size (trajectories) & $32{,}000$ \\
Sampling temperature & $1.0$ \\
GRPO group size $G$ (rollouts per $(Q,A)$) & $1024$ \\
Number of training $(Q,A)$ pairs $B$ & $500$ \\
Number of validation $(Q,A)$ pairs & $200$ \\
Maximum episode length $L_{\max}$ & $12$ \\
Reward & $+1$ if answer reached, $0$ otherwise \\
Total training epochs & $1{,}000$ \\
Random seed & $42$ (single seed; no multi-seed averaging) \\
\bottomrule
\end{tabular}
\end{table}

Each training epoch performs the same loop: for every one of the $B=500$ training pairs, $G=1024$ trajectories are rolled out under the current policy; rewards are normalised within each group (group baseline); $K=4$ inner GRPO updates are then performed on the resulting batch; finally the validation set is evaluated. The training curve is reported in Fig.~1b of the main text.

\subsection{PSV Equivalence Relation: Operational Definition}
\label{app:slidingpuzzle:psv}

The PSV equivalence relation of Eq.~(1) of the main text is implemented exactly via subtree hashing rather than via a continuous distance.

For every active decision point $(Q,h)$ we expand its complete future tree, pruning any branch whose conditional next-action probability is below $p_{\text{prune}}=0.05$ (such branches contribute negligibly to behaviour). At every surviving node we compute the rounded next-action distribution
\begin{equation}
\widetilde{\pi}_\theta(\cdot\mid Q,h) \;=\; \big(\,\mathrm{round}_r(\pi_\theta(a_1\mid Q,h)),\ \ldots,\ \mathrm{round}_r(\pi_\theta(a_K\mid Q,h))\big),
\end{equation}
where $\mathrm{round}_r(\cdot)$ rounds to $r$ decimal places. The signature of $(Q,h)$ is the recursively constructed tuple
\begin{equation}
\sigma(Q,h) \;=\; \big(\,\widetilde{\pi}_\theta(\cdot\mid Q,h),\ \sigma(Q,h\!\cdot\! a_1),\ \ldots,\ \sigma(Q,h\!\cdot\! a_K)\big),
\end{equation}
which is a Python tuple equipped with exact equality. Two histories are declared equivalent if and only if their signatures are byte-identical: $(Q,h)\sim_r (Q',h') \iff \sigma(Q,h)=\sigma(Q',h')$. Equivalently, this enforces $|\pi_\theta(\cdot\mid Q,h)-\pi_\theta(\cdot\mid Q',h')|\le 5\times 10^{-r-1}$ at every shared decision point in the future tree, so the rounding precision $r$ acts as the discrete counterpart of the threshold $\epsilon$ in Eq.~(1) of the main text. The robustness check below uses $r\in\{1,2,3,4\}$.

The set of \emph{active decision points} is constructed by enumerating, for every $(Q,A)$ pair in the union of the training set and the validation set, the BFS subtree $\mathcal{T}_Q$ rooted at $Q$ to depth $L_{\max}$, retaining only those $(Q,h)$ pairs for which (i) $h$ has not yet hit the answer or the depth limit and (ii) every action in $h$ has conditional probability $\ge p_{\text{prune}}=0.05$. The size of the active set is then taken as the \emph{sum} (not the deduplication) of node counts across all problems:
\begin{equation}
|\mathcal{S}_{\text{active}}| \;=\; \sum_{(Q,A)\in\,\text{train}\cup\text{val}} \big|\mathcal{T}_Q^{\text{active}}\big|.
\end{equation}
Crucially, the signature $\sigma$ above does \emph{not} include $Q$ as part of its hash --- it depends only on the local future-tree structure of the policy. Consequently, two histories from \emph{different} problems whose future subtrees are byte-identical contribute the same signature and are merged into a single PSV class. This implementation choice is what makes $R_c$ sensitive to the cross-problem state sharing emphasised in §EFFECTIVE MARKOV REPRESENTATION of the main text: as RLVR carves shared reasoning paths across problems, the count of distinct signatures $|\mathcal{S}_{\text{PSV}}|$ grows much more slowly than $|\mathcal{S}_{\text{active}}|$, driving $R_c$ upward in Stage 2.

\subsection{Compression Ratio $R_c$}
\label{app:slidingpuzzle:rc}

The compression ratio is defined as
\begin{equation}
R_c \;=\; 1 - \frac{|\mathcal{S}_{\text{PSV}}|}{|\mathcal{S}_{\text{active}}|},
\end{equation}
where $|\mathcal{S}_{\text{active}}|$ is the active-point count of §S\ref{app:slidingpuzzle:psv} (summed across all train$\cup$val problems) and $|\mathcal{S}_{\text{PSV}}|$ is the number of distinct subtree signatures $\sigma(\cdot)$ obtained by globally deduplicating signatures across the same union. By construction $R_c\in[0,1)$: $R_c=0$ when every active history is its own equivalence class (no compression), and $R_c\to 1$ when many distinct histories collapse onto a small set of effective states (high compression).

The trajectory of $R_c$ during training (Fig.~1b of the main text) traces a non-monotonic V-shape: an initial drop driven by policy diversification, followed by a rebound and saturation in a high-compression regime. The V-shape is the direct empirical signature of PSV-class refinement followed by cross-problem state sharing, both of which are predictions of the effective Markov representation introduced in §EFFECTIVE MARKOV REPRESENTATION of the main text.

\begin{figure}[!htbp]
\centering
\includegraphics[width=\columnwidth]{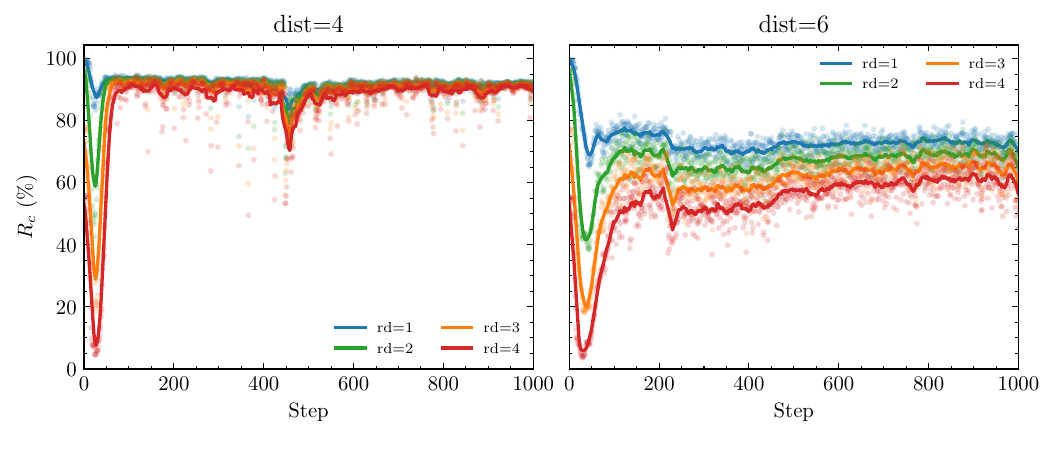}
\caption{\textbf{Robustness of the empirical PSV criterion to rounding precision.} In the dist$=4$ (left) and dist$=6$ (right) experiments, $R_c$ is computed after rounding the policy probabilities to $r\in\{1,2,3,4\}$ decimal places. All four curves exhibit the same V-shaped dynamics and differ only by an overall vertical shift: the looser the rounding (rd$=1$), the easier the equivalence criterion is to satisfy and the higher $R_c$. At the strictest setting rd$=4$, the post-training compression ratio is $R_c\approx 0.93$ (dist$=4$) and $R_c\approx 0.84$ (dist$=6$), with late-training fluctuations spanning roughly $78$--$93\%$, confirming that the compression is a genuine structural feature of the policy rather than a rounding artefact.}
\label{fig:rc_rounddigits}
\end{figure}

\subsection{Robustness of $R_c$ to Rounding Precision}
\label{app:slidingpuzzle:robustness}

The threshold $\epsilon$ in the PSV equivalence relation is implemented by the rounding precision $r$. Fig.~\ref{fig:rc_rounddigits} reports the $R_c$ trajectories at $r\in\{1,2,3,4\}$ for both dist$=4$ and dist$=6$ training conditions. All four curves exhibit the same V-shaped dynamics and differ only by an overall vertical shift: the looser the rounding (rd$=1$), the easier the equivalence criterion is to satisfy and the higher $R_c$.

At the strictest setting rd$=4$, the post-training compression ratio is $R_c\approx 0.93$ for dist$=4$ and $R_c\approx 0.84$ for dist$=6$ (final-epoch values $0.925$ and $0.842$, last-10-epoch means $0.94$ and $0.825$, peaks $0.964$ and $0.872$; data from \texttt{rc\_psv\_rd4.json} of the two experiments). Across the late training window the values fluctuate roughly between $78\%$ and $93\%$. The fact that $R_c$ remains in this range at the strictest threshold confirms that the compression is a genuine structural feature of the policy rather than an artefact of the rounding precision.

\section{From Empirical Reasoning Graphs to Predictive State Variables}
\label{app:coweb}

\begin{figure}[!htbp]
\centering
\includegraphics[width=0.8\columnwidth]{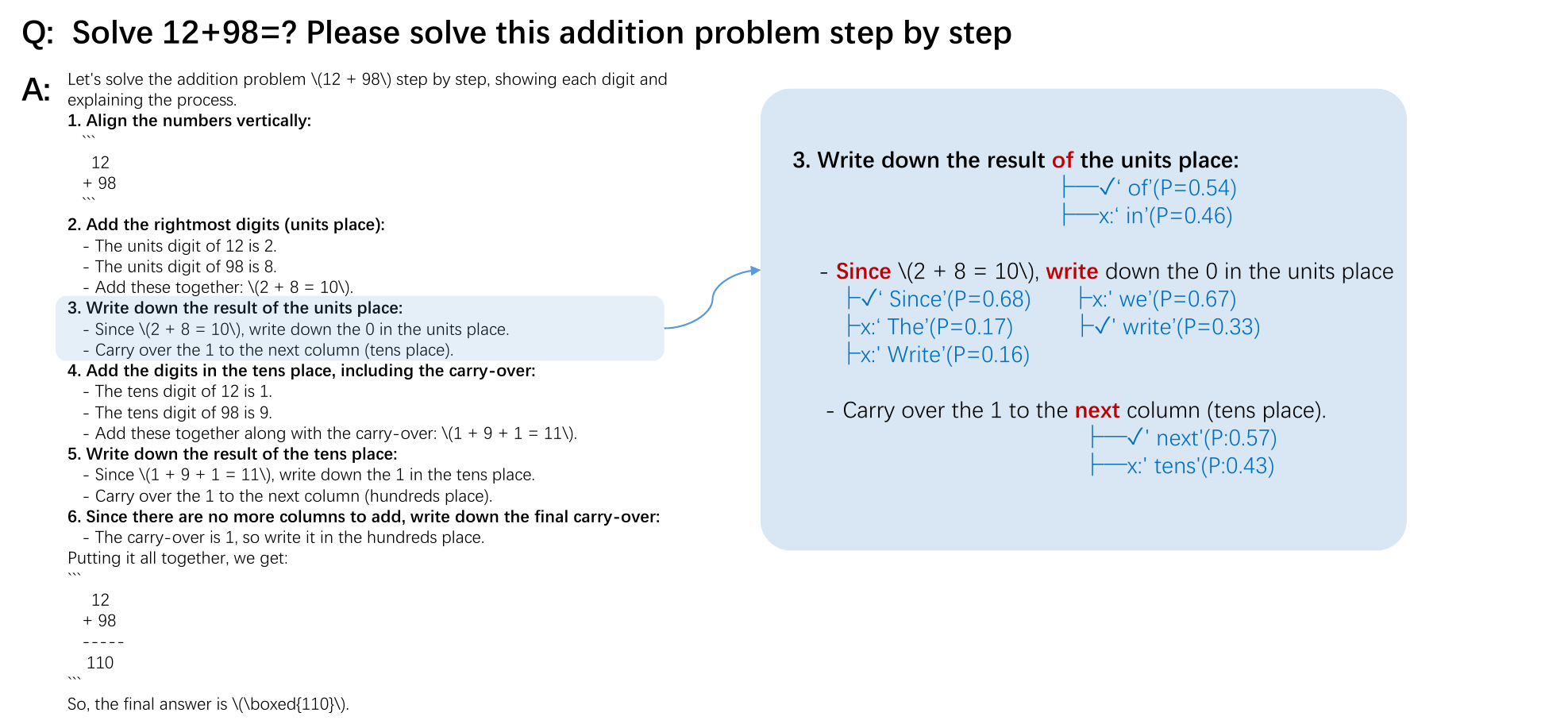}
\caption{Illustration of the LLM reasoning process. A chain-of-thought is composed of stable, low-entropy text chunks (concepts), connected by high-entropy ``forking tokens'' that represent decision points (e.g., ``next'' vs.\ ``tens''). The probabilities at these junctures represent transition weights on an underlying reasoning graph. Figure adapted from Ref.~\citep{cai2025learning}.}
\label{fig:cot_example}
\end{figure}

A growing body of work represents LLM reasoning as a graph of effective states linked by transitions. Empirically, reasoning chains are decomposed into low-entropy ``concept'' chunks separated by high-entropy ``forking tokens'' at which the model commits to one of several semantically distinct continuations (Fig.~\ref{fig:cot_example}). Recent constructions take this view further: \citep{minegishi2025topology} clusters hidden-state representations of segmented reasoning steps to build task-specific graphs and reports that more capable models develop graphs of larger diameter; \citep{wang2024understandingreasoningabilitylanguage} analyses LLM reasoning through reasoning-path aggregation during pretraining.

Useful as these constructions are, post-hoc graph extraction faces a fundamental \emph{node-stability problem}. K-means cluster centroids over hidden states do not have an intrinsic dynamical meaning: the resulting graph shifts when the representation layer, segmentation granularity, clustering algorithm, or training checkpoint changes. The mapping from low-level token sequences to high-level semantic concepts is many-to-one and itself evolves during RLVR training, so an extracted graph cannot reliably track the same nodes across training time.

The effective Markov representation introduced in the main text resolves this difficulty in a principled way. Predictive state variables (PSVs) define nodes as equivalence classes of histories whose future conditional distributions agree to a fixed precision (Eq.~1 of the main text). This definition is layer-agnostic and clustering-free: two histories belong to the same PSV class if and only if they generate indistinguishable futures, an intrinsic property of the model's policy. Finite parametric capacity bounds the number of distinguishable conditional distributions by $\exp(O(d))$, so a great many distinct histories must necessarily collapse onto a small set of effective states. The sliding-puzzle experiment in §EFFECTIVE MARKOV REPRESENTATION of the main text directly verifies this compression: a 14K-parameter Transformer trained with GRPO achieves a compression ratio of 78--93\% across rounding precisions, confirming that the PSV graph is dynamics-defined and observable rather than hypothesised.

\begin{figure}[!htbp]
\centering
\includegraphics[width=0.8\columnwidth]{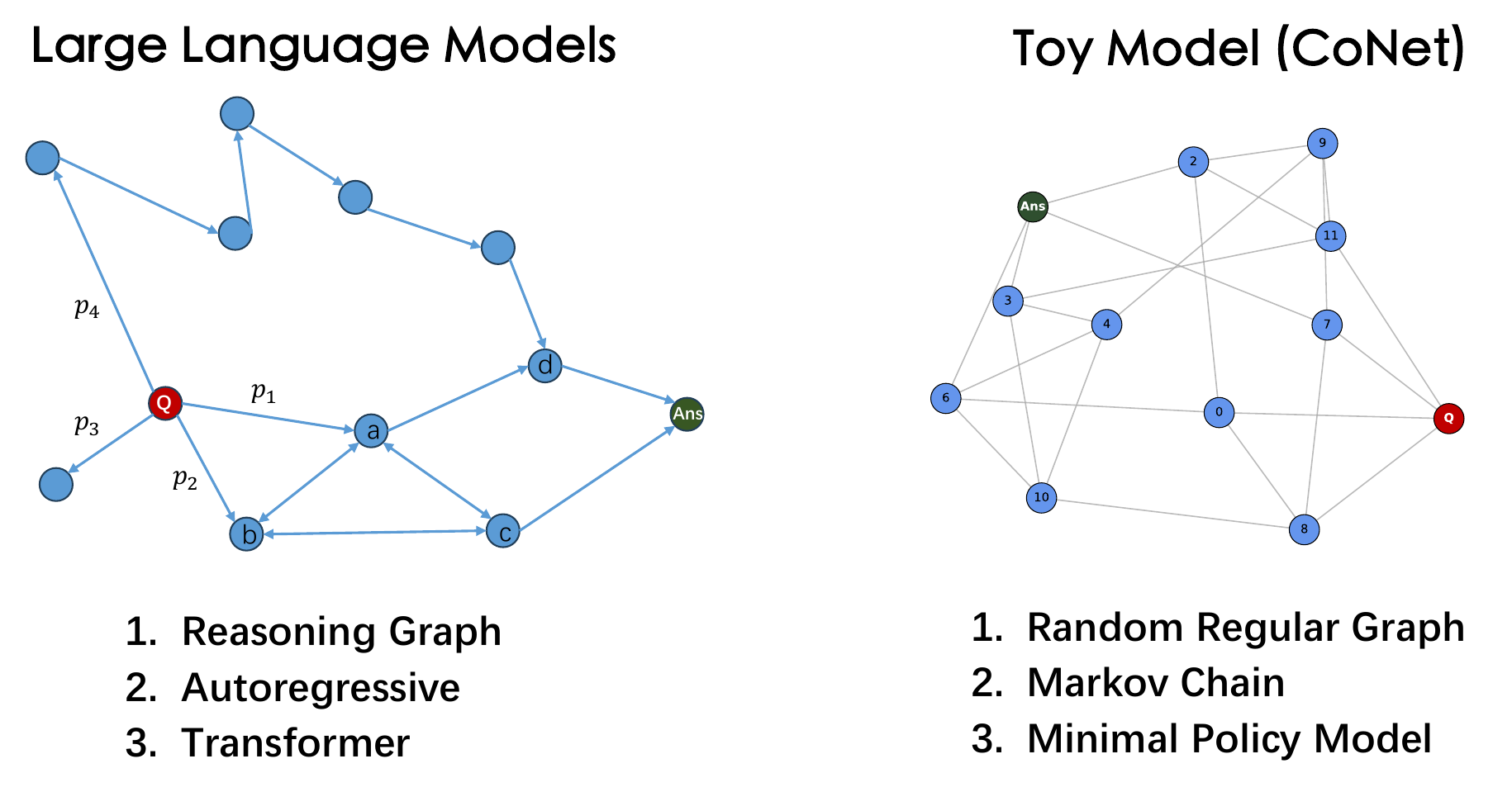}
\caption{\textbf{From an implicit PSV graph to a minimal CoNet model.} The reasoning process in an LLM (left) is a traversal on the implicit PSV equivalence-class graph; transition probabilities are determined by the autoregressive Transformer policy. CoNet (right) is the minimal explicit counterpart: a fixed $K$-regular random graph with a learnable Markov-chain policy. Both share the structure ``random walk on a sparse graph''; CoNet exposes the policy parameters directly so that merging and frustration can be observed. Figure adapted from Ref.~\citep{cai2025learning}.}
\label{fig:conet_abstraction}
\end{figure}

The CoNet picture used in §THE CONCEPT NETWORK is the abstraction obtained by replacing the implicit, dynamics-defined PSV graph with an explicit, parametrically-controlled random graph (Fig.~\ref{fig:conet_abstraction}). It is a minimal computational platform on which the same merging+frustration mechanisms can be studied with full microscopic observability, complementing the indirect PSV measurement on the sliding-puzzle Transformer.

\section{CoNet: Origin and Multi-Task Extension}
\label{app:CoNet-dynamics}

CoNet was introduced by \citep{cai2025learning} as a minimal abstraction of LLM reasoning dynamics. The model represents the concept space as a $K$-regular random graph, and a reasoning chain as a policy-guided random walk; the transition policy $\pi_\theta(j\mid i)\propto\theta_{ij}$ is governed by learnable parameters $\theta_{ij}$ representing the connection strength from node $i$ to node $j$. This policy is updated by GRPO on the success or failure of a trajectory from question $Q$ to answer $A$. In the original single-task setting, CoNet captured the Learning-at-Criticality (LaC) phenomenon---a sharp, sigmoidal learning transition accompanying generalisation peaks under extremely sparse training data---and identified it with a continuous learning phase transition.

The methodological contribution of the present work is to deploy CoNet in a multi-task regime, with many concurrent question--answer pairs sharing a single set of transition parameters. The shift is conceptually decisive. With one task, the policy only needs to find an efficient path; with many tasks sharing parameters, distinct reasoning paths inevitably collide on shared nodes and place mutually exclusive demands on the same outgoing-edge probabilities, which is the microscopic origin of the frustration mechanism analysed in §TWO CORE MECHANISMS of the main text. The single-task LaC sigmoid (monotonic length decrease + sigmoidal reward) is recovered as a limiting case in §S\ref{app:single_task_dynamics}.

In the cross-scale comparison of the main text (Fig.~3 of the main text), CoNet is the smallest of three systems---CoNet, a 14K-parameter sliding-puzzle Transformer, and the 1.5-billion-parameter DSR-1.5B---that all display the same two-stage dynamics and V-shaped path-length curve. CoNet's role is therefore not to ``bypass'' a problem unique to LLMs but to provide an analytically transparent counterpart: every transition probability is logged in full and every connected component of high-probability edges is directly visualisable, enabling the microscopic tests of merging and frustration reported in the main text.

A single-step hop on CoNet's graph represents fast, intuitive ``System 1'' association; a multi-step traversal guided by the learned policy represents deliberate, compositional ``System 2'' reasoning. Importantly, CoNet does not model the Transformer's parameters themselves: it models the emergent semantic network induced by inference---the structure that connects questions to answers through chains of effective states. The hyperparameters of the multi-task CoNet experiments reported here are detailed in §S\ref{appendix:conet_implementation}.

\section{CoNet Implementation Details}
\label{appendix:conet_implementation}

We configured CoNet with a specific set of structural and learning parameters. The underlying concept space was instantiated as a directed random regular graph with $N=800$ nodes, where each node has a uniform out-degree of $k=40$. In this graph, we established a multi-task learning environment consisting of $128$ randomly generated Q-A node pairs. The transition parameters $\theta_{ij}$ for all edges were initialized randomly. The use of a multi-task setting with 128 concurrent Q-A pairs is the fundamental mechanism that induces structural competition. While a single task would only require finding one efficient path, the presence of numerous tasks provides competing reinforcement signals. When multiple Q-A paths need to traverse a common node, they may ``disagree'' on the optimal subsequent step, leading to conflicting policy updates. This microscopic conflict is the origin of the macroscopic ``frustration'' discussed in the main text, compelling the system to find a shared, globally coherent structure that can arbitrate these conflicts, rather than settling for 128 disconnected ``skill islands''.

The policy was optimized using a variant of the GRPO algorithm (\citep{shao_deepseekmath_2024, drgrpo, yu_dapo_2025}). For a given Q-A pair, $n_{\text{rollout}} = 128$ reasoning paths (indexed by $m$) were sampled, with each path's exploration capped at a maximum length of 20 steps. Each path received a reward based on its success, and its advantage $A_m$ (relative to the average reward over all 128 rollouts) was used to guide the update of the policy parameters $\theta_{ij}$. The update rule is given by:
$$\Delta\theta_{ij} \propto \sum_{m} A_m \nabla_{\theta_{ij}} \log \pi_\theta(j|i)$$
This advantage-based update functions as a competitive reinforcement mechanism, strengthening above-average paths and suppressing underperforming ones. This ``winner-take-more'' dynamic acts as an algorithmic sculpting tool, systematically pruning the vast majority of the initial $k=40$ edges from each node to carve out the sparse, efficient backbone of the concept web. The learning rate was set to $0.04$ to control the speed and stability of this training process.

Crucially, the two-stage learning dynamic is not an artifact of fine-tuned parameters but a robust emergent property of the model. Repeated experiments show that this behavior---a fast-learning phase followed by a slow integration plateau, and the corresponding V-shaped response length---consistently emerges under a broad condition: when the number of Q-A pairs, $N_{QA}$, is in the regime of $1 \ll N_{QA} \ll N$. This dynamic is qualitatively different from the single Q-A pair ($N_{QA}=1$) case, which produces a simple sigmoidal learning transition (the LaC phenomenon). The condition is essential: if $N_{QA}$ is too small, there is no systemic competition to drive the formation of a unified web; if it is too large, the conflicting reinforcement signals prevent any coherent structure from forming. The robust emergence of these LLM-like signatures in this intermediate regime confirms that our CoNet setup successfully reproduces the essential dynamics of structural self-organization under multi-task reinforcement.



\section{Structural Evolution of the Concept Web in CoNet}
\begin{figure}[!htbp]
\centering
\begin{subfigure}[b]{0.8\textwidth}
    \centering
    \includegraphics[width=\linewidth]{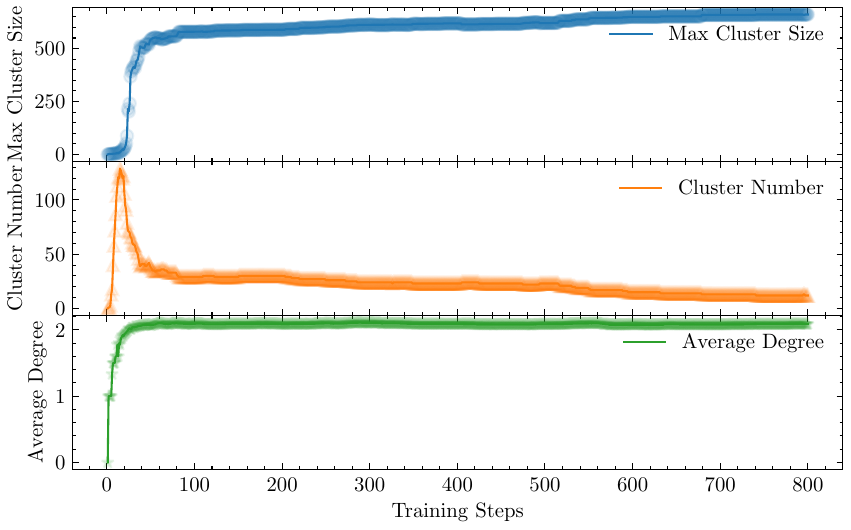}
\end{subfigure}

\caption{
\textbf{The Formation of the Concept Web in CoNet.} This figure illustrates the evolution of network clusters and topology during training, showing a structural reorganization from isolated skills to a unified conceptual web. In the initial phase, the Cluster Number (orange curve) spikes, indicating the rapid discovery of numerous disconnected ``skill islands''. The peak marks a critical structural juncture. Following this peak, the Cluster Number steadily declines as these islands begin to integrate. This merging process is confirmed by the corresponding monotonic growth in the Max Cluster Size (blue curve), which represents the formation of a single giant component. Crucially, the Average Degree of the conept web (green curve) rapidly converges to and remains pinned at $\approx 2$ throughout the slow-learning phase. This persistent sparsity confirms that the global integration occurs within a strictly constrained, tree-like topology, validating our central hypothesis.}
\label{fig:conet_property}
\end{figure}

This section provides direct, quantitative evidence from the CoNet model for the two-stage learning dynamic---a fast acquisition process followed by a slow integration phase---that underlies our theory. Fig.~\ref{fig:conet_property} serves as the central exhibit for this analysis, tracking three key structural properties over the course of training: the total number of disconnected solution clusters (the ``skill islands''), the size of the single largest cluster, namely the concept web and the average degree of the concept web.

The first stage of training ( $\sim$ 0-50 steps), corresponding to the fast-learning phase, is characterized by the rapid proliferation of ``skill islands''. The quantitative evidence for this is the steep, initial rise of the ``Cluster Number'' (orange curve) in Fig.~\ref{fig:conet_property}. Each new cluster represents the discovery of a distinct, high-confidence reasoning path for a specific problem. During this phase, the system operates in a greedy, parallel fashion, finding many efficient, local solutions without regard for their interconnection.

The shift to the slow-learning phase occurs at the critical juncture marked by the peak of the orange curve around step 20. At this point, the system has exhausted the ``low-hanging fruit'' of simple, isolated solutions and has reached a state of maximal fragmentation. This peak signals the onset of the ``maximally frustrated state'' discussed in the main text. The subsequent, gradual decline of the ``Cluster Number'' curve is the direct signature of this frustrated state, representing the arduous and competitive process of integrating existing islands rather than discovering new ones. The ``Max Cluster Size'' (blue curve), which begins its steep, sustained ascent at precisely this moment, confirms this shift from local discovery to a global integration objective.

The slow-learning phase is thus defined by the coalescence of these islands into a unified concept web. The plot reveals the two complementary trends that define this stage: a steady decline in the total number of clusters (orange curve) and a simultaneous, relentless growth in the size of the largest cluster (blue curve). The causal link is direct: the largest component grows precisely by absorbing smaller, independent clusters, which necessarily reduces the total cluster count. Moreover, the average degree of the concept web remains pinned at $\approx2$ throughout the expansion, indicating that the concept web maintains a strictly sparse, tree-like topology. This structural evolution culminates in a state where a single, giant component---the functional concept web---dominates the network. This process provides a direct, mechanical explanation for the V-shaped response length signature: Stage 1 shortens paths by finding local solutions, while Stage 2 lengthens them by forcing the policy to traverse longer, connective paths between previously distant concepts.

\begin{figure}[!htbp]
\centering
\begin{subfigure}[b]{0.31\textwidth}
    \centering
    \includegraphics[width=\linewidth]{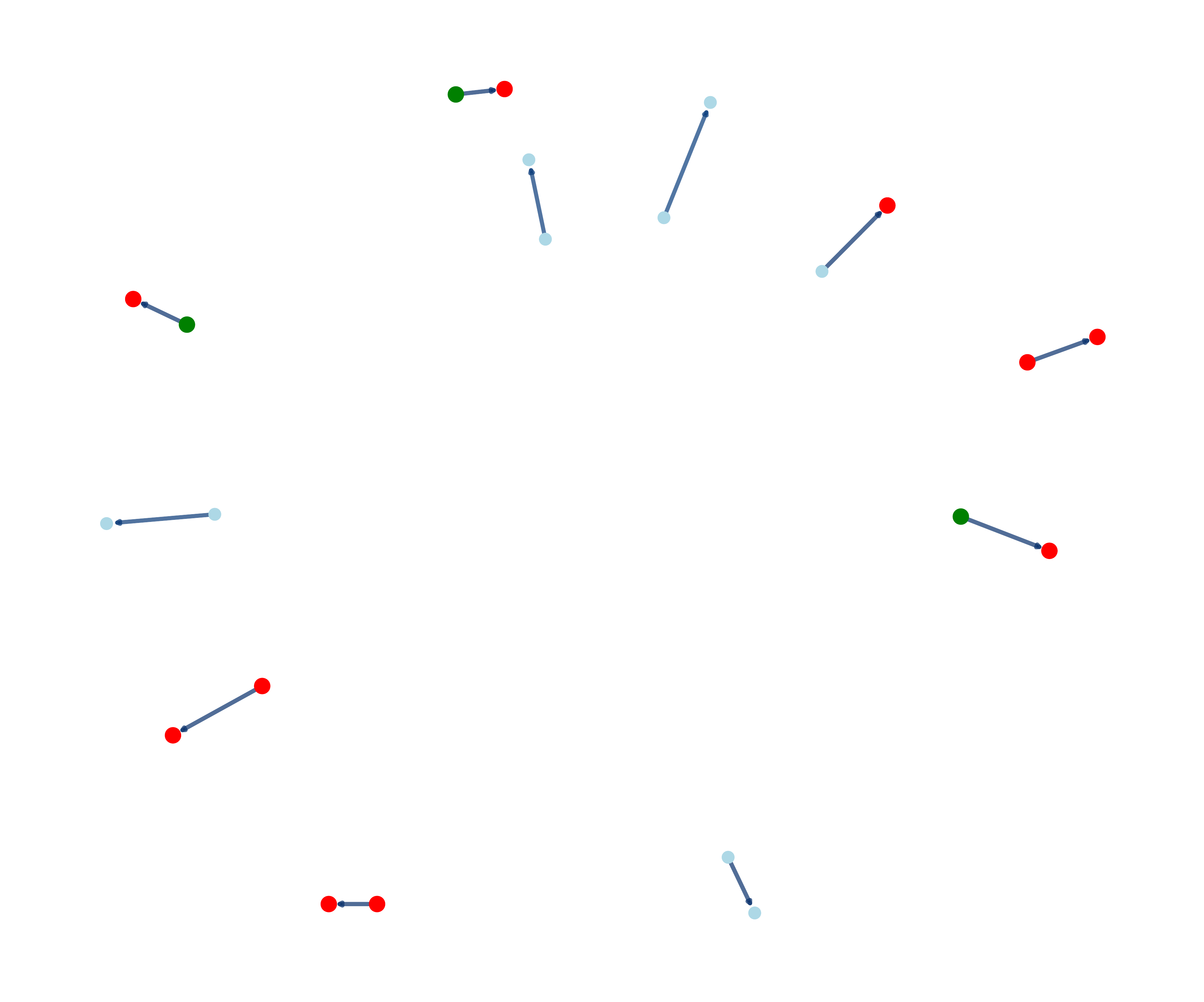}
    \caption{Step 20}
    \label{fig:net_step20}
\end{subfigure}
\hfill
\begin{subfigure}[b]{0.31\textwidth}
    \centering
    \includegraphics[width=\linewidth]{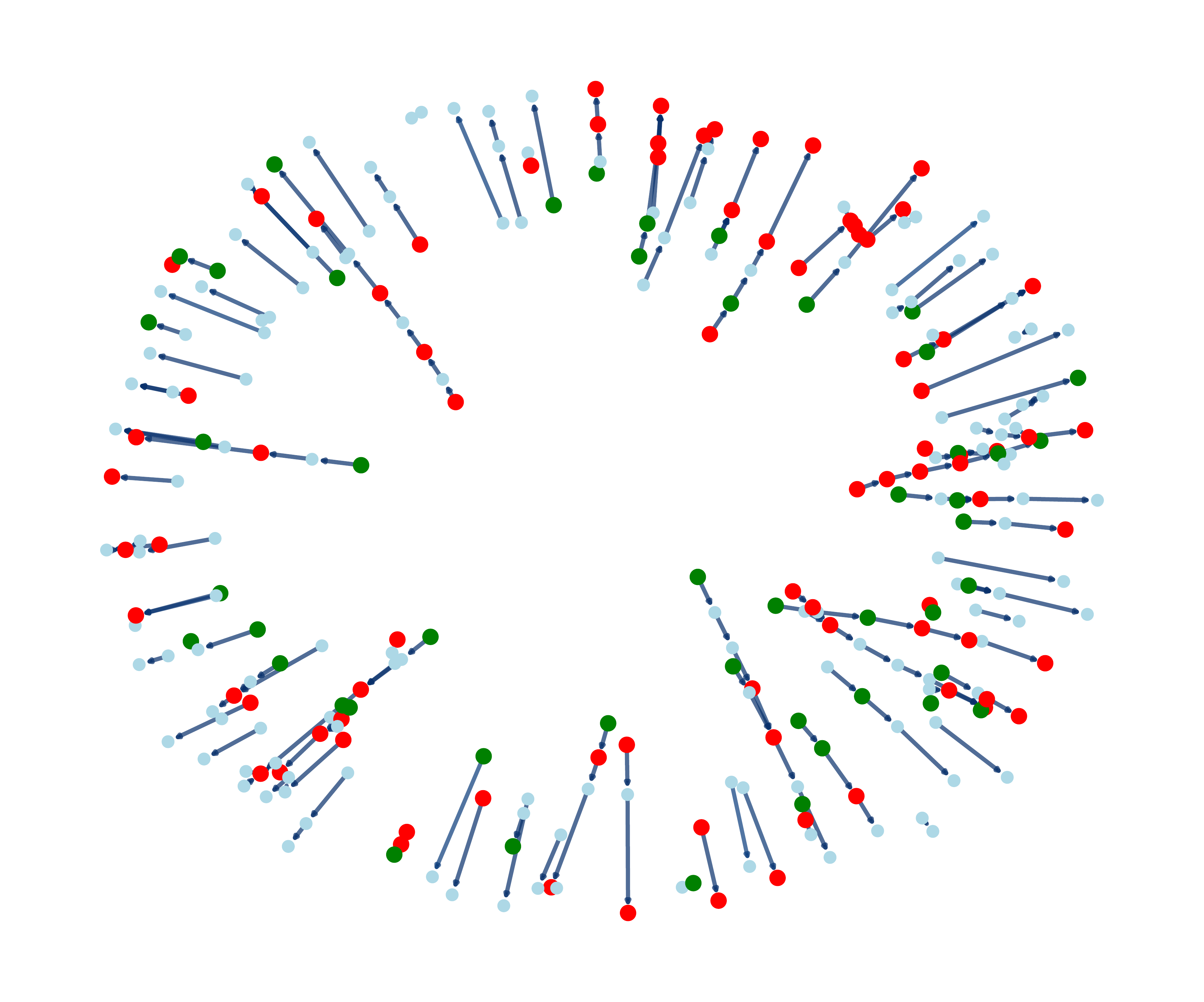}
    \caption{Step 50}
    \label{fig:net_step50}
\end{subfigure}
\hfill
\begin{subfigure}[b]{0.31\textwidth}
    \centering
    \includegraphics[width=\linewidth]{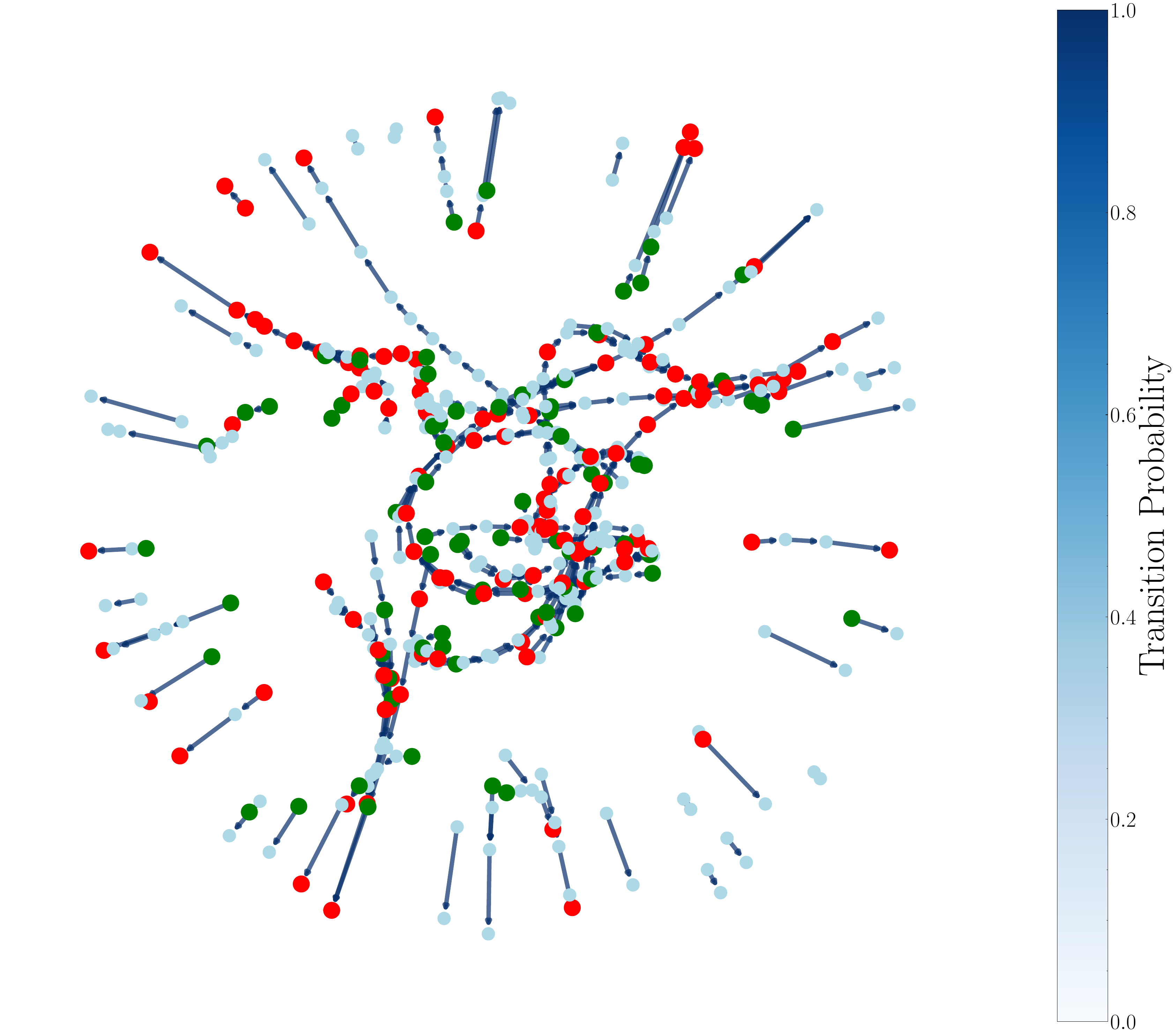}
    \caption{Step 800}
    \label{fig:net_step800}
\end{subfigure}
\caption{\textbf{Direct visualization of inverse-tree freezing in CoNet.} Network snapshots at the three training steps marked on Fig.~\ref{fig:conet_property}, retaining only edges with transition probability $p>0.95$. Green = question nodes, red = answer nodes, blue = intermediate nodes; edge colour encodes the transition probability (right colour bar). Unlike Fig.~4 of the main text, which extracts only the largest inverse tree at each step, this figure shows \emph{all} surviving high-probability edges, exposing the full landscape of connected components throughout training. (\textbf{a}) Step~20: multiple short, disjoint subtrees nucleate independently across the graph---the nascent ``skill islands''---coinciding with the peak of the cluster-number curve in Fig.~\ref{fig:conet_property} and the onset of the maximally frustrated state. (\textbf{b}) Step~50: a dominant component has emerged through the merging of compatible paths at shared nodes, while a handful of small subtrees have yet to be absorbed; this is the regime in which frustration-induced forgetting is most active. (\textbf{c}) Step~800: the surviving high-probability edges have coalesced into a single connected inverse tree with average degree $\langle k\rangle\approx 2$, the frozen-in concept web that supports the trained policy.}
\label{fig:conet_micro}
\end{figure}

These three macroscopic indicators (Fig.~\ref{fig:conet_property}) are made microscopically concrete by the network snapshots in Fig.~\ref{fig:conet_micro}, which visualize the same three training steps directly. The Step~20 snapshot makes the ``proliferation of skill islands'' literal: many short, disjoint paths populate the graph in parallel, exactly as expected when nucleation dominates and merging has not yet set in. The Step~50 snapshot shows the structural fingerprint of the maximally frustrated state: a dominant connected component has begun to grow by welding compatible paths at shared nodes, but several subordinate islands persist, sustaining the inter-task competition that drives frustration-induced forgetting. By Step~800, the high-probability edge set has frozen into a single, sparse inverse tree---the concept web---whose multi-input, single-output topology and average degree $\langle k\rangle\approx 2$ are precisely the structural ingredients that the main text identifies as responsible for both the V-shaped response-length signature and the bridge-node fragility under SFT.

\section{Geometric Origin of the Rising Branch of Response Length}
\label{app:length-geometry}

The main text argues that the lengthening of correct responses during Stage~2 is a direct geometric consequence of the inverse tree: with average degree $\langle k\rangle\approx 2$ and out-degree close to $1$ at almost every node, the network has no short-range shortcuts, so as the inverse tree extends to cover more question--answer pairs the typical question-to-answer geodesic grows monotonically. This appendix provides the corresponding microscopic evidence in CoNet: tracking the largest inverse tree alongside the distribution of correct path lengths during Stage~2 shows that the two evolve together, exactly as the inverse-tree picture predicts.

\begin{figure}[!htbp]
\centering
\begin{subfigure}[t]{0.42\textwidth}
    \centering
    \includegraphics[height=0.55\linewidth]{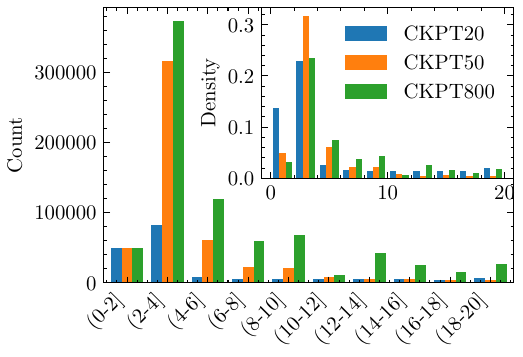}
    \caption{Correct-response length distribution}
    \label{fig:cluster_response_histogram}
\end{subfigure}
\hfill
\begin{subfigure}[t]{0.18\textwidth}
    \centering
    \includegraphics[width=0.95\linewidth]{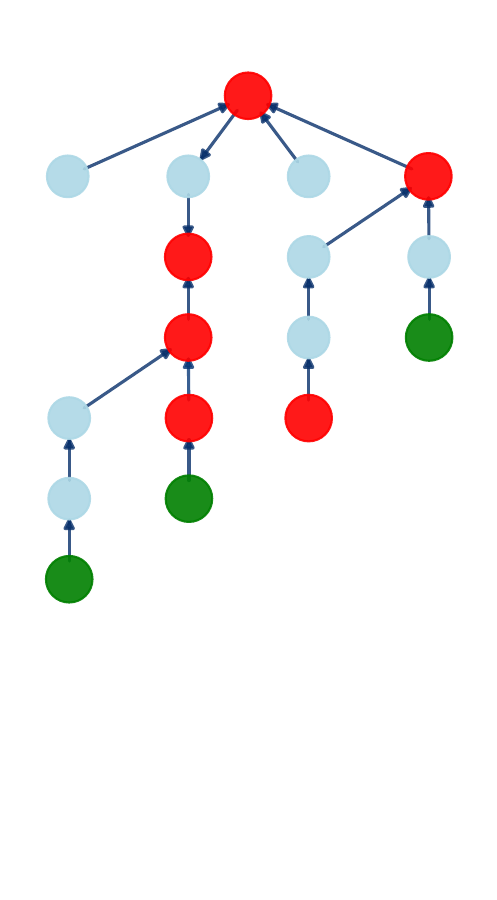}
    \caption{Step 50}
    \label{fig:cluster_step50}
\end{subfigure}
\hfill
\begin{subfigure}[t]{0.34\textwidth}
    \centering
    \includegraphics[height=4cm]{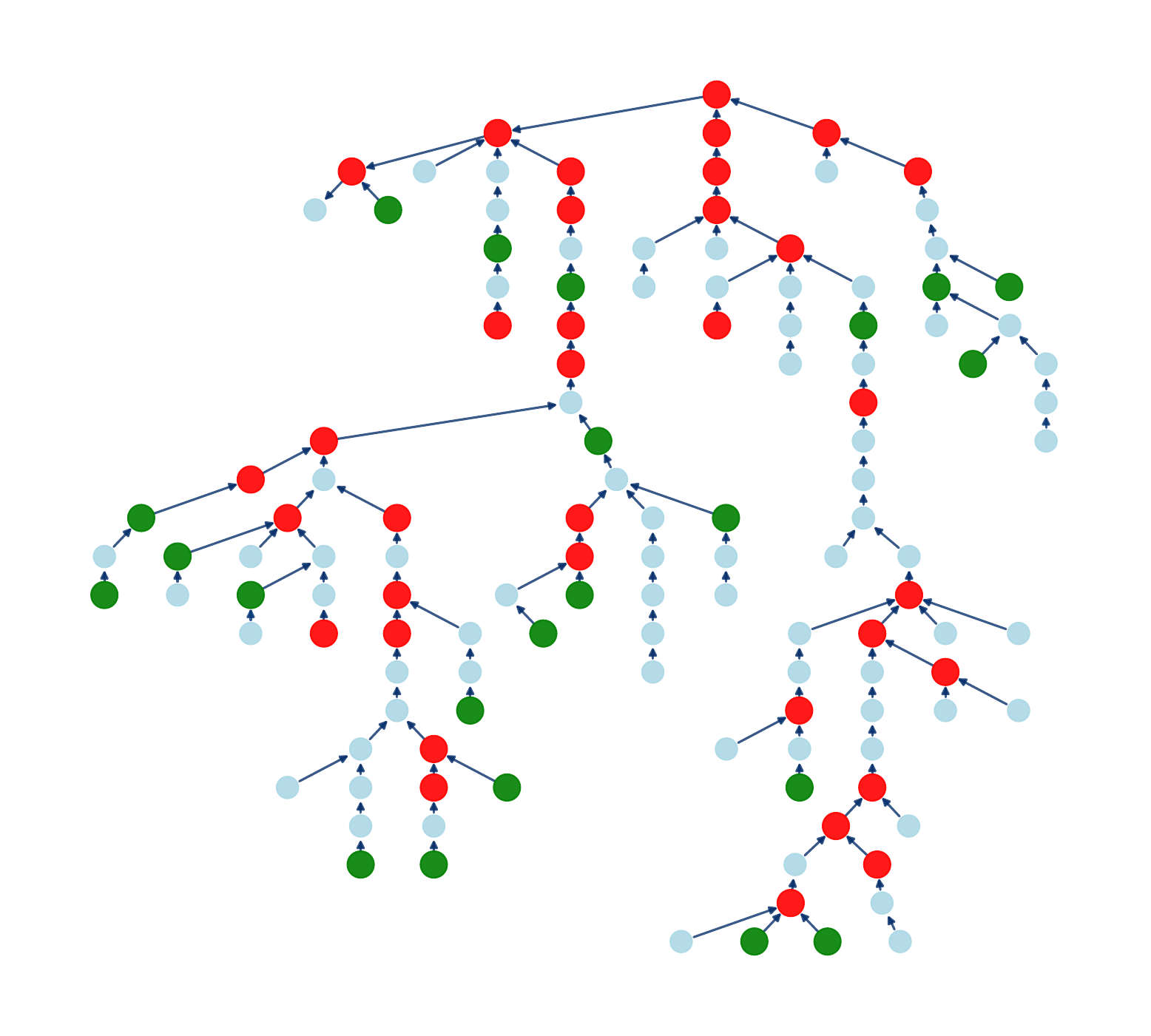}
    \caption{Step 800}
    \label{fig:cluster_step800}
\end{subfigure}
\caption{\textbf{Inverse-tree growth and the rising branch of response length in CoNet.} (\textbf{b, c}) Largest connected component at $p>0.95$ at Step~50 and Step~800, drawn with the same colour and marker scheme as Fig.~\ref{fig:conet_micro}. The inverse tree expands substantially between the two steps---absorbing additional question--answer pairs through merging at shared nodes---while keeping its sparse, multi-input/single-output character: average degree stays $\langle k\rangle\approx 2$ (cf.\ Fig.~\ref{fig:conet_property}). (\textbf{a}) Distribution of correct solution path lengths over the same training window (raw counts, with the probability density shown in the inset). As the inverse tree grows, the length distribution shifts decisively to the right; both the mean and the upper tail move outward, while the support remains bounded above by the maximum episode length.}
\label{fig:conet_web_length}
\end{figure}

The rightward shift of the length distribution in Fig.~\ref{fig:conet_web_length}a is not an incidental feature of CoNet's hyperparameters but the direct macroscopic signature of the geometric constraint imposed by the frozen inverse tree. Once Stage~2 begins, new question--answer pairs are no longer mastered by discovering fresh short-cut paths in the underlying random graph; they are mastered by being grafted, through merging, onto the existing inverse tree at one of its branching nodes. Each such grafting attaches an additional segment of length set by the graph distance from the new question node to the nearest node already on the tree, and this distance is itself bounded below by the tree's growing radius. As a result, the typical solution path length must increase together with the size of the inverse tree, exactly as observed when the largest component expands from Step~50 (Fig.~\ref{fig:conet_web_length}b) to Step~800 (Fig.~\ref{fig:conet_web_length}c).

This sparse-tree-geodesic argument also clarifies what the rising branch is \emph{not}. It is not driven by the model adopting longer reasoning styles, nor by a switch to harder problems, nor by stochastic noise on already-mastered paths: in CoNet none of these can vary, since the graph, the problem set, and the policy parameterisation are all fixed by construction. The only quantity that changes between Step~50 and Step~800 is the topology of the high-probability sub-network, and yet the response-length histogram shifts decisively. The rising branch of the V is therefore a structural readout of inverse-tree growth, in line with the corresponding mechanism stated in the main text. Self-correction or reflection-style cycles are fully compatible with this picture: they are sparse, local loops on the same backbone, occasional rather than dense, and do not appreciably change the global average degree from $\langle k\rangle\approx 2$.

\section{Learning by Phase Transition in the Slow-Learning Stage}
\label{app:stage2-phase-transition}

\begin{figure}[!htbp]
\centering
\includegraphics[width=0.8\textwidth]{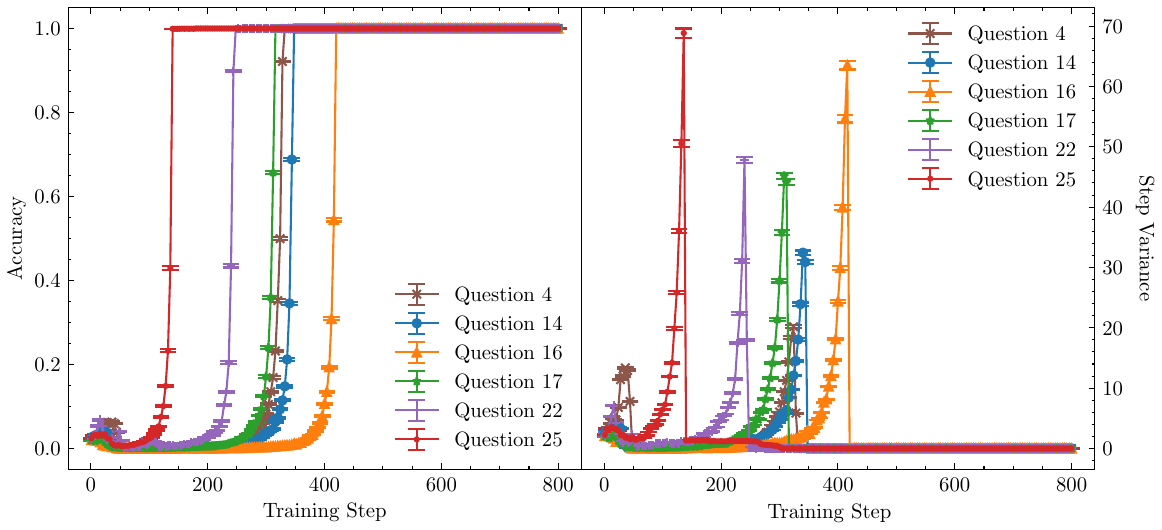}
\caption{\textbf{Per-Problem Critical Dynamics During the Slow-Learning Stage.} These plots reveal the microscopic learning dynamics for a representative subset of problems that successfully learn during the slow-learning stage, providing evidence for localized phase transitions. \textbf{(Left)} Accuracy trajectories for these problems. All problems shown exhibit a sharp, sigmoidal-like increase in accuracy rate in the training process, characteristic of a critical learning event. \textbf{(Right)} Corresponding variance of the solution path length. A sharp, pronounced peak in variance directly coincides with the rapid accuracy gain for each problem. This dual signature is the hallmark of a continuous phase transition, suggesting that the slow-learning stage in CoNet is composed of discrete, sharp learning events rather than a uniform, gradual process.}
\label{fig:per_problem_critical_dynamics}
\end{figure}

In the main text, we argue that new knowledge acquisition during the slow-learning stage is not gradual but occurs via discrete, sharp events. This appendix provides detailed, per-problem evidence for this learning by phase transition mechanism, drawing a direct analogy to critical phenomena in statistical physics.

Fig. \ref{fig:per_problem_critical_dynamics} isolates a representative subset of problems mastered late in the training process. The first signature of a phase transition is visible in the left panel: each problem exhibits a sharp, sigmoidal-like jump in accuracy. In the language of statistical physics, accuracy acts as an order parameter, signaling the system's transition from a disordered (unsolved) state to an ordered (solved) state. An abrupt change in the order parameter is a primary indicator of a phase transition.

The second, more telling signature is the pronounced, transient peak in the variance of the solution path length, shown in the right panel. This provides a deeper link to the physics of critical phenomena. In physical systems, the variance (or fluctuations) of an order parameter is related to the system's susceptibility---its response sensitivity to external perturbations. At the critical point of a continuous phase transition, this susceptibility is known to diverge, creating a characteristic sharp peak. A famous example of this is the lambda ($\lambda$) peak observed in the specific heat of liquid helium at the normal-to-superfluid transition. The variance peaks in our figure are the direct analogue of this phenomenon. This ``dual signature''---an abrupt rise in the order parameter (accuracy) and a lambda-like peak in its fluctuations (variance)---is the classic hallmark of a critical learning transition. This same analysis was performed in the original work on LaC, where the variance peak in the single-task CoNet was identified as a key hallmark of a learning phase transition~\citep{cai2025learning}. 

The intuitive reason for this variance peak is straightforward. When accuracy is very low, nearly all sampled paths fail, resulting in low variance. When accuracy is very high, the model has converged on a dominant, efficient reasoning path, again leading to low variance. It is only during the narrow transition window that a rich diversity of paths coexists---some still failing, while others explore newly viable routes to the solution. This temporary coexistence of competing strategies is what produces the characteristic peak in variance.

Taken together, these results provide strong microscopic validation for our claims. They show that the slow-learning stage in CoNet is not a process of uniform, gradual improvement. Instead, it is punctuated by a series of discrete, critical learning events, where individual skills are integrated into the concept web one by one via localized phase transitions at the frontier.

\section{The SFT and Annealed-RLVR Algorithm in CoNet}
\label{app:CoNet-SFT}
This appendix details the implementation and validation of the Annealed-RLVR algorithm in CoNet, our ``computational microscope'' for LLM reasoning dynamics. The algorithm is designed to intervene at the point of ``maximum frustration'' with SFT to resolve the competitive bottlenecks that arise during the formation of the sparse concept web.

\subsection{SFT Implementation in CoNet}
CoNet's minimalistic policy model significantly simplifies the standard SFT process. This process involves first using the original model's policy to sample a successful reasoning path for a given problem. Then, for this selected path, we adjust its transition probabilities. For any transition on the path with a probability below a certain threshold, we raise it to that threshold value. To ensure that the probability distribution of each node remains valid, the probabilities of other outgoing transitions are proportionally reduced. The value of this target transition probability is a tunable parameter that determines the intervention's effect. For our annealing strategy, we set this value to 0.1. To induce catastrophic forgetting, we use a more aggressive value of 0.5.
\subsection{Expeimental Results}
\begin{figure}[!htbp] 
    \centering 
    \includegraphics[width=\textwidth]{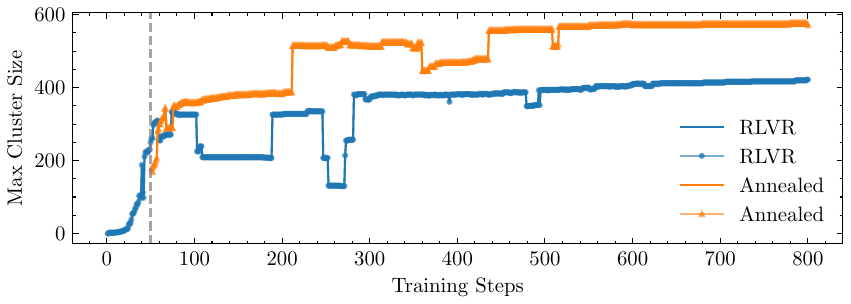} \caption{\textbf{Structural Impact of Annealed-RLVR on the Concept Web.} Evolution of concept web for standard RLVR (blue) versus Annealed-RLVR (orange). The SFT intervention at step 50 (dashed line) induces an immediate drop in the Annealed model's cluster size. Subsequently, the Annealed model recovers and surpasses the standard RLVR baseline, which exhibits slower growth, ultimately forming a larger final concept web.} 
    \label{fig:conet_cluster_comparison}
\end{figure}
\begin{figure}[!htbp]
    \centering
    \includegraphics[height=0.39\textwidth]{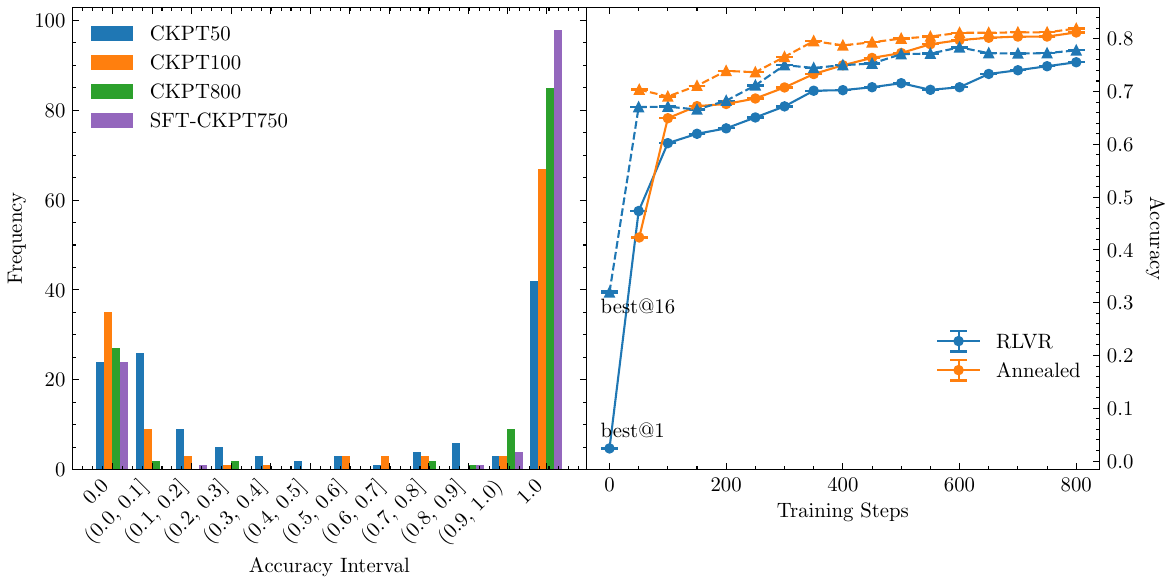}
    \caption{\textbf{Validation of Annealed-RLVR in CoNet.} 
\textbf{(Left)} The histogram of per-problem accuracy shows that the annealed model (SFT-CKPT750) significantly reduces the number of completely unsolved problems (Accuracy = 0.0) and increases the number of mastered problems (Accuracy = 1.0) compared to the standard model (CKPT800). 
\textbf{(Right)} The line plot of training dynamics shows that immediately after the SFT intervention at step 50, best@1 accuracy dips while best@16 accuracy sharply increases, indicating a successful trade of exploitation for exploration. The final annealed policy (``Annealed'') ultimately achieves superior performance over the baseline (``RLVR'').}
\label{fig:conet_annealing_results}
\end{figure}

The algorithm consists of a strategically timed SFT phase, followed by a resumption of standard RLVR training. The SFT intervention is applied at training step 50, the empirically determined state of maximum frustration, marking the crossover from the formation of disconnected ``skill islands'' to their subsequent integration. At this stage, we first identify all problems with an accuracy below 0.1, select a total of 50 low-accuracy problems, and sample a known successful reasoning path for each. This targeted adjustment serves as the ``heating'' step in our simulated annealing analogy; it gently increases the discoverability of latent correct paths to restore the system's exploratory capacity. Following this brief SFT, standard RLVR training is resumed, which functions as the ``cooling'' phase, allowing the now more exploratory policy to settle into a new, more globally optimal configuration.

\change{}{Fig. \ref{fig:conet_cluster_comparison} provides the direct structural evidence for this annealing process. As shown by the orange ``Annealed'' curve, the SFT intervention at step 50 locally disrupted network connections, resulting in a decline in the concept web size. However, this disruption is precisely what allows the system to overcome the integration bottleneck. While the standard RLVR model (blue curve) is shown to grow slowly, trapped in the maximally frustrated state, the Annealed model's cluster size rapidly recovers and grows to a larger final scale. This demonstrates that the SFT intervention, by resolving the competitive frustration, enables the subsequent RLVR ``cooling'' phase to successfully expand the network and integrate a greater number of ``skill islands''.}

The success of this SFT-RLVR cycle is also confirmed by the results in Fig. \ref{fig:conet_annealing_results}. The line plot on the right shows the immediate effect of the intervention: best@1 accuracy (a proxy for exploitation) temporarily dips, while best@16 accuracy (a proxy for exploration) surges. This demonstrates a successful trade of greedy exploitation for enhanced exploration, and the long-term benefit is clear, as the final annealed policy significantly surpasses the baseline RLVR policy. The histogram on the left reveals the underlying mechanism for this improvement. The SFT-RLVR cycle successfully resolves the frustration bottleneck by reducing the population of completely unsolved problems (Accuracy = 0.0) and correspondingly increasing the number of fully mastered problems (Accuracy = 1.0) compared to the standard model.

In summary, the CoNet experiment provides strong evidence for our theory. By applying a targeted and conservative SFT intervention at the point of maximum frustration, the Annealed-RLVR algorithm effectively resolves the system's competitive bottlenecks, enabling it to discover a more robust and better reasoning policy.

\section{LLM Training Details}
\label{app:training}
\subsection{Base Model and GRPO Setup}
We used the \textbf{DeepSeek-R1-Distill-Qwen-1.5B} model~\citep{deepseek-r1}, trained on the DeepScaleR-Preview-Dataset~\citep{deepscaler2025} with their modified version of the veRL open-source engine~\citep{sheng2025hybridflow,verl_github}. Our Group Relative Policy Optimization ({GRPO}) hyperparameters followed the first-stage settings, which specified a response length of 8192, as outlined in the DeepScaleR paper~\citep{deepscaler2025}.

\subsection{SFT Dataset for Annealed-RLVR}
\label{app:sft_annealed}
The SFT dataset for the \textbf{annealed-RLVR} experiment (labeled ``Annealed'' in Fig.~6 of the main text) was constructed to focus on difficult problems. First, we trained the base model using GRPO for 100 steps to create a checkpoint (CKPT100), which is in the vicinity of the maximally frustrated state. We then used this checkpoint to generate 8 rollouts for each of the approximately 40,000 problems in the training set. For the roughly 10,000 problems that yielded no correct responses, we generated an additional 52 rollouts, bringing the total to 60 per problem. From this data, we created the SFT dataset by selecting all correct responses for problems where the initial model accuracy was below 10\%, which yielded \textbf{3,933 trajectories}. Finally, we fine-tuned the model on these trajectories for two epochs with a learning rate of $3\times 10^{-5}$. The resulting model served as the new starting point for continuing GRPO training up to step 1,700.

\subsection{SFT Dataset for Catastrophic Forgetting}
The SFT dataset for the \textbf{catastrophic forgetting} experiment (labeled ``SFT+RLVR'' in Fig.~5b of the main text) was designed to specifically target problems forgotten by a more saturated model, using a data generation process that involved multiple checkpoints. The process began with problem filtering: we used a checkpoint from step 600 (CKPT600, deep within the slow-learning stage) to generate 8 rollouts for all training problems. We then identified the subset of problems with no correct responses and generated another 32 rollouts for them, again using CKPT600. Problems that still had no correct responses were selected as the final ``forgotten'' set. For the second stage of targeted data generation, each problem in this forgotten set was used to generate a total of 96 rollouts: 32 from the \textbf{base model}, 32 from CKPT100, and 32 from CKPT600. We then collected correct responses from problems that had at least 3 correct answers among these 96 rollouts, resulting in \textbf{798 trajectories}. This SFT dataset was used to fine-tune the model for two epochs with a learning rate of $5\times 10^{-5}$ before resuming GRPO training.

\subsection{Computational Cost Analysis}
\label{app:cost}
\change{}{The overall computational complexity of \textbf{Annealed-RLVR} is dominated by the iterative {GRPO} training, making it nearly identical to the standard RLVR baseline. The only addition is a \textbf{one-time cost} for the SFT intervention, which is applied once at the point of maximal frustration (around step 100 in our case). This one-time cost consists of two phases:
\begin{enumerate}
    \item \textbf{Evaluation Pass:} An evaluation pass over the $N$ problems in the training set (where $N \approx 40,000$) is required to identify the low-accuracy problems. As described in Section~\ref{app:sft_annealed}, this involved generating 8 to 60 rollouts per problem. This cost can be formally described as $O(N \cdot k)$ for a fixed $k$ rollouts, or $O(N \cdot \log(k))$ if using an adaptive, binary-like rollout strategy to find problems below the accuracy threshold.
\item \textbf{SFT Phase:} The SFT itself is performed on the small subset of resulting correct trajectories. Letting $\eta$ be the accuracy threshold parameter (e.g., $\eta=0.1$ for our 10\% threshold), the SFT cost is proportional to $O(\eta N)$. As detailed in Section~\ref{app:sft_annealed}, this amounted to only \textbf{3,933 trajectories} (approximately $\eta N$) in our experiment.
\end{enumerate}
This $O(\eta N)$ SFT cost is significantly smaller than the evaluation cost. Both of these marginal, one-time costs are much smaller than the total iterative cost of the full GRPO training experiment (e.g., 1,700 steps).
}

\section{LLM Evaluation Details}
\label{app:eval}
We evaluated checkpoints from our RLVR (``GRPO'') and annealed-RLVR (``Annealed'') training experiments on two datasets: 512 randomly selected training problems and the Minerva math dataset~\citep{minerva1,lewkowycz2022solving}. The evaluation parameters were kept consistent with the training configuration (temperature: $0.6$, top-p: $1.0$, response length: $8192$, etc.). From the $512$ rollout results, we calculated the best@k accuracy and the standard error. Fig.~6 of the main text illustrates the full performance curves, while Table~\ref{tab:accuracy_summary_error} highlights the results from representative checkpoints.

\begin{figure}[!htbp]
    \centering
    \begin{subfigure}{0.32\textwidth}
        \centering
        \includegraphics[width=\linewidth]{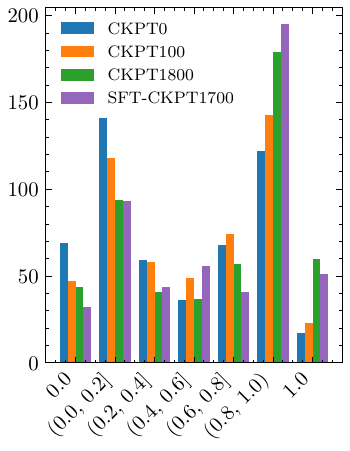}
        \caption{Training Sample (512 Problems)}
        \label{fig:dsrsft_minerva_hist}
    \end{subfigure}
    \begin{subfigure}{0.32\textwidth}
        \centering
        \includegraphics[width=\linewidth]{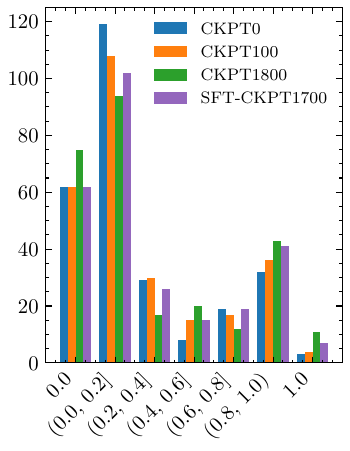}
        \caption{Minerva}
        \label{fig:dsrsft_minerva_hist}
    \end{subfigure}
    \begin{subfigure}{0.32\textwidth}
        \centering
        \includegraphics[width=\linewidth]{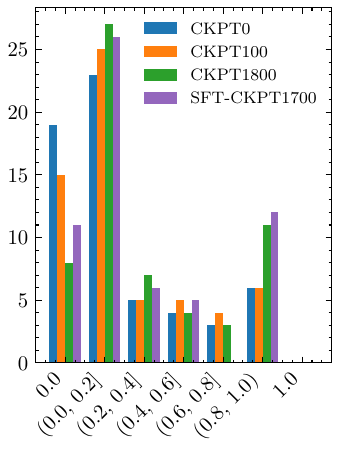}
        \caption{AIME 2024/2025}
        \label{fig:dsrsft_aime2425_hist}
    \end{subfigure}
    \caption{\textbf{Mechanism of Annealed-RLVR: Accuracy Distributions.} This figure complements the best@k curves in the main text by revealing the underlying mechanism of performance improvement. It shows the per-problem accuracy histograms for the checkpoints of Annealed-RLVR ("Annealed") versus the standard RLVR ("RLVR") baseline, evaluated on (a) the in-distribution set (Random 512), (b) the OOD Minerva dataset, and (c) the OOD AIME 2024/2025 datasets. The histograms demonstrate that the Annealed-RLVR intervention systematically reduces the population of unsolved problems (low-accuracy bins) and increases the population of mastered problems (high-accuracy bins).}
    \label{fig:annealed_rlvr_results_hist}
\end{figure}

\begin{table}[htbp]
\centering
\caption{Best@k Accuracy (\%) with Standard Error at Key Training Checkpoints}
\label{tab:accuracy_summary_error}
\begin{tabular}{llccccccc}
\toprule
& & \multicolumn{7}{c}{\textbf{Checkpoints (Steps)}} \\
\cmidrule(lr){3-9}
\textbf{Metric} & \textbf{Method} & \textbf{0} & \textbf{100} & \textbf{200} & \textbf{600} & \textbf{1000} & \textbf{1400} & \textbf{1800} \\
\midrule
\multicolumn{9}{c}{\textbf{Dataset: Randomly Selected 512 Training Problems}} \\
\midrule
\multirow{2}{*}{best@1}  & GRPO & 43.66(6) & 50.17(6) & 52.45(6) & 56.49(6) & 58.00(6) & 58.90(6) & {59.55(6)} \\
                         & Annealed & ---      & 50.00(6) & 52.27(6) & 55.80(6) & 58.10(6) & 59.46(6) & \textbf{60.31(6)} \\
\cmidrule(lr){2-9}
\multirow{2}{*}{best@4}  & GRPO & 61.32(8) & 68.46(8) & 70.38(8) & 72.99(7) & 73.46(7) & 73.71(7) & {74.20(7)} \\
                         & Annealed & ---      & 69.49(8) & 71.51(8) & 73.75(7) & 75.10(7) & 75.91(7) & \textbf{76.77(7)} \\
\cmidrule(lr){2-9}
\multirow{2}{*}{best@16} & GRPO & 72.97(12) & 78.60(11) & 80.12(11) & 81.60(10) & 81.61(10) & 81.62(10) & {81.86(10)} \\
                         & Annealed & ---       & 80.62(11) & 82.22(11) & 83.15(10) & 83.77(10) & 84.48(10) & \textbf{85.28(10)} \\
\cmidrule(lr){2-9}
\multirow{2}{*}{best@128}& GRPO & 82.16(24) & 86.98(23) & 87.81(20) & 87.47(19) & {88.23(21)} & 87.77(19) & 87.99(22) \\
                         & Annealed & ---       & 88.00(19) & 89.35(20) & 89.37(19) & 89.98(18) & 90.23(18) & \textbf{91.22(18)} \\
\midrule
\multicolumn{9}{c}{\textbf{Dataset: Minerva}} \\
\midrule
\multirow{2}{*}{best@1}  & GRPO & 23.85(7) & 26.08(7) & 26.83(7) & 27.63(7) & 28.53(7) & 28.45(7) & {28.77(7)} \\
                         & Annealed & ---      & 26.09(7) & 26.62(7) & 27.73(7) & 28.28(7) & 28.75(7) & \textbf{28.88(7)} \\
\cmidrule(lr){2-9}
\multirow{2}{*}{best@4}  & GRPO & 37.84(12) & 40.77(12) & 41.33(12) & 41.12(12) & {41.46(11)} & 41.04(11) & 41.32(11) \\
                         & Annealed & ---       & 40.63(12) & 40.90(12) & 41.66(12) & 42.04(11) & \textbf{42.53(11)} & 42.39(11) \\
\cmidrule(lr){2-9}
\multirow{2}{*}{best@16} & GRPO & 51.39(21) & \textbf{54.29(21)} & 54.20(21) & 53.05(20) & 52.38(19) & 52.12(20) & 51.95(19) \\
                         & Annealed & ---       & 53.80(21) & 53.71(21) & 53.74(20) & 53.86(21) & {54.24(21)} & 53.50(20) \\
\cmidrule(lr){2-9}
\multirow{2}{*}{best@128}& GRPO & 66.6(5) & 68.8(4) & \textbf{69.8(5)} & 66.2(4) & 65.8(5) & 64.9(4) & 64.4(4) \\
                         & Annealed & ---       & 68.2(5) & 68.3(4) & 68.4(5) & 68.0(4) & {69.4(5)} & 67.8(5) \\
\midrule
\multicolumn{9}{c}{\textbf{Dataset: AIME 2024 \& 2025}} \\
\midrule
\multirow{2}{*}{best@1}  & GRPO & 20.11(0.10) & 23.08(0.11) & 24.04(0.12) & 26.98(0.12) & 27.77(0.11) & 28.11(0.11) & {28.10(0.11)} \\
                         & Annealed & ---      & 23.27(0.11 & 23.86(0.12 & 26.77(0.12 & 27.67(0.11 & \textbf{28.26(0.11} & 27.93(0.11 \\
\cmidrule(lr){2-9}
\multirow{2}{*}{best@4}  & GRPO & 33.31(0.17) & 38.44(0.19) & 40.33(0.19) & 42.74(0.18) & 43.36(0.19) & 43.51(0.18) & \textbf{43.88(0.18)} \\
                         & Annealed & ---       & 39.36(0.19) & 40.21(0.19) & 43.38(0.19) & 43.76(0.19) & {43.79(0.19)} & 43.42(0.19) \\
\cmidrule(lr){2-9}
\multirow{2}{*}{best@16} & GRPO & 45.3(0.3) & 52.4(0.3) & 54.4(0.3) & 55.5(0.3) & 56.9(0.3) & 56.1(0.3) & {57.1(0.3)} \\
                         & Annealed & ---       & 54.1(0.3) & 54.7(0.3) & 57.6(0.3) & 57.7(0.3) & \textbf{58.1(0.3)} & {57.8(0.3)} \\
\cmidrule(lr){2-9}
\multirow{2}{*}{best@64} & GRPO & 55.9(0.5) & 62.4(0.5) & 64.2(0.5) & 64.5(0.6) & 65.9(0.6) & 64.8(0.6) & {66.2(0.6)} \\
                         & Annealed & ---       & 63.8(0.5) & 64.2(0.5) & 67.0(0.5) & 66.7(0.5) & 67.5(0.5) & \textbf{67.7(0.5)} \\
\bottomrule
\end{tabular}
\end{table}
\newpage
\section{Supplementary Evidence: Single-Task Dynamics}
\label{app:single_task_dynamics}

\begin{figure}[!htbp]
\centering
\begin{subfigure}[b]{0.45\textwidth}
    \centering
    \includegraphics[width=\linewidth]{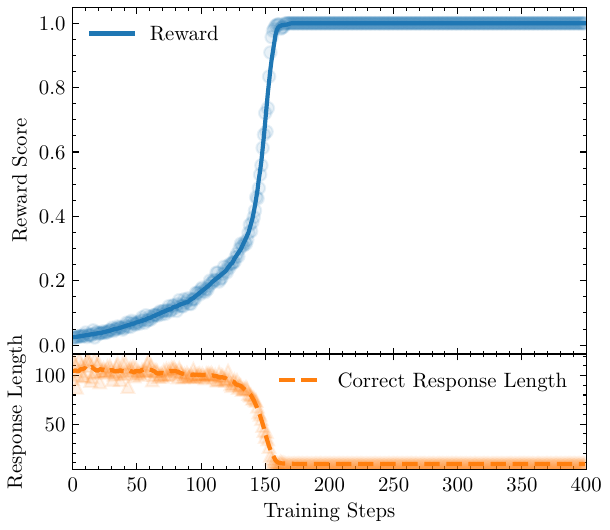}
    \caption{CoNet}
    \label{fig:dynamics_conet_single}
\end{subfigure}
\hfill 
\begin{subfigure}[b]{0.45\textwidth}
    \centering
    \includegraphics[width=\linewidth]{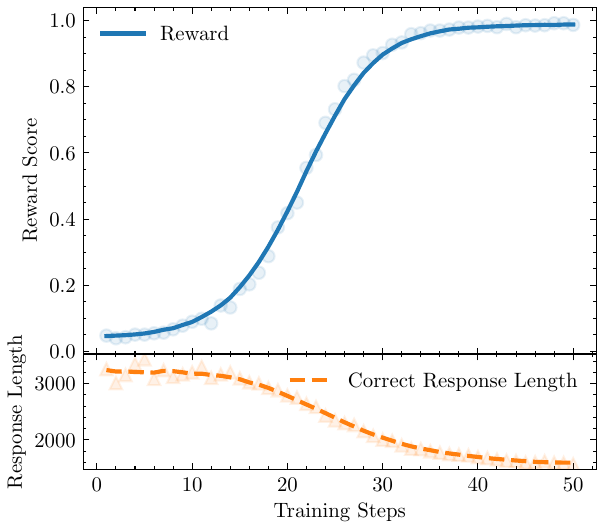}
    \caption{LLM}
    \label{fig:dynamics_deepscaler_single}
\end{subfigure}

\caption{\textbf{Single-Task Dynamics} We compare the training dynamics of CoNet (a) and the DeepScaleR-1.5B LLM (b) when trained on a single problem. The single-task experiment exhibits: (i) The reward curves (top panels) follow a sigmoidal, phase-transition-like trajectory; (ii) The correct response length (bottom panels) decreases monotonically, converging to the efficient paths. Lines in (a) and (b) are smoothed with a 5-step and 4-step moving average respectively.}
\label{fig:single_task_comparison}
\end{figure}

We reproduced the single-problem training experiment studied by \citep{cai2025learning} using the DeepScaleR protocol~\citep{deepscaler2025}. Specifically, we selected a single problem from the DeepScaleR-Preview-Dataset~\citep{deepscaler2025} and repeated the single-task training experiment with their modified version of the veRL (Volcano Engine Reinforcement Learning). As shown in Fig.~\ref{fig:single_task_comparison}, the single-task dynamics exhibit a sigmoidal learning curve and a monotonically decreasing response length, standing in sharp contrast to the two-stage reward dynamic and V-shaped response length observed during multi-task training. CoNet faithfully reproduces these distinct behaviors when restricted to the same single-task condition. 

This single-task training more effectively highlights the phase-transition-like learning mechanism described in the main text. By restricting the system to a single task, we isolate the fundamental phase-transition dynamics of skill acquisition: the model focuses solely on identifying an optimal trajectory among multiple feasible paths. Without the structural requirement to integrate conflicting skills into a shared, sparse backbone, the system avoids the ``maximally frustrated state.'' Consequently, the response length decreases monotonically as the policy converges to a fixed optimal path, and the reward curve follows a smooth, sigmoidal trajectory characteristic of a continuous phase transition, free from competitive frustration. We also note that the reward curve is influenced by finite-size effects imposed by the maximum response length constraint, this phenomenon remains a key subject for further theoretical investigation. 

This capacity to replicate the distinct phenomenologies of both conditions---from the phase transition of isolated learning to the competitive frustration inherent in multi-trajectory integration---provides strong evidence that CoNet serves as a minimal model for investigating the emergent mechanisms of RLVR.

\end{document}